\documentclass[11pt]{article}

\usepackage[final]{acl}
\usepackage{adjustbox}
\usepackage{pdflscape}
\usepackage{times}
\usepackage[T1]{fontenc}
\usepackage[utf8]{inputenc}
\usepackage{microtype}
\usepackage{inconsolata}
\usepackage{graphicx}
\usepackage{booktabs}
\usepackage{longtable}
\usepackage{array}
\usepackage{xcolor}
\usepackage{tabularx}

\usepackage{tikz}
\usetikzlibrary{calc,positioning}
\newcommand{\taxcite}[2]{\citep{#1}}
\newcommand\blfootnote[1]{\begingroup\renewcommand\thefootnote{}\footnote{#1}\addtocounter{footnote}{-1}\endgroup}
\title{Beyond Textual Chain-of-Thought: \\
A Survey on Action-Grounded Reasoning in Autonomous Driving}

\author{Zhengxu Tang\textsuperscript{*}, Xiaozhou Zhang\textsuperscript{*}, Guofeng Cui, Ziyu Gong, Zi Wang, \\
  \bfseries Yunfei Shi, Ruifeng Deng, Chengzhi Qi, Ke Chen, Sachin Patil, \\
  \bfseries Tianjun Xiao, Langechuan Liu, Pichao Wang\textsuperscript{\dag} \\
  NVIDIA \\}

\begin{document}
\maketitle
\blfootnote{\textsuperscript{*}\,Equal Contribution.}
\blfootnote{\textsuperscript{\dag}\,Corresponding Author.}

\begin{abstract}

Chain-of-thought (CoT) reasoning powers generative models by eliciting intermediate steps before producing an answer. In autonomous driving, the answer is a continuous action. Thus its reasoning must share the same spatiotemporal structure as the physical world. This survey studies the resulting shift from textual CoT to action-grounded reasoning. Surveying 171 papers, including 130 method papers and 41 benchmarks, datasets, surveys, and analysis papers, we propose a representation-centered taxonomy that treats the form of the intermediate state as the organizing axis. We systematize the 130 methods into four categories: language-based, visual-spatial, latent-dynamic, and externalized reasoning, further divided into 13 subtypes tied to distinct regions of interests. Our synthesis shows that the open frontier of reasoning in driving agents lies in intermediate representations that can be grounded in the real world, coupled to real-time action, and verified under safety-critical systems. Project page:  \url{https://github.com/tangzhengxu/awesome-av-cot}.
\end{abstract}

\section{Introduction}
\label{sec:intro}

\looseness=-1 Autonomous driving has cycled through several architectural generations, all driven by the same pressure: better generalization. As autonomous driving systems increasingly incorporate large language models (LLMs), vision-language models (VLMs), and vision-language-action (VLA) models for scene understanding, decision explanation, and action generation~\citep{hwang2024emma,tian2024drivevlm}, they aim to escape the local minimum of pure end-to-end driving through explicit thinking traces, made possible by stronger backbones pretrained on large-scale multimodal data. But what should a driving model think before it acts? We need a reasoning layer flexible enough to bridge scene interpretation and trajectory generation, yet explicit enough to be supervised and audited.

\begin{figure}[t]
  \centering
  \includegraphics[width=\columnwidth]{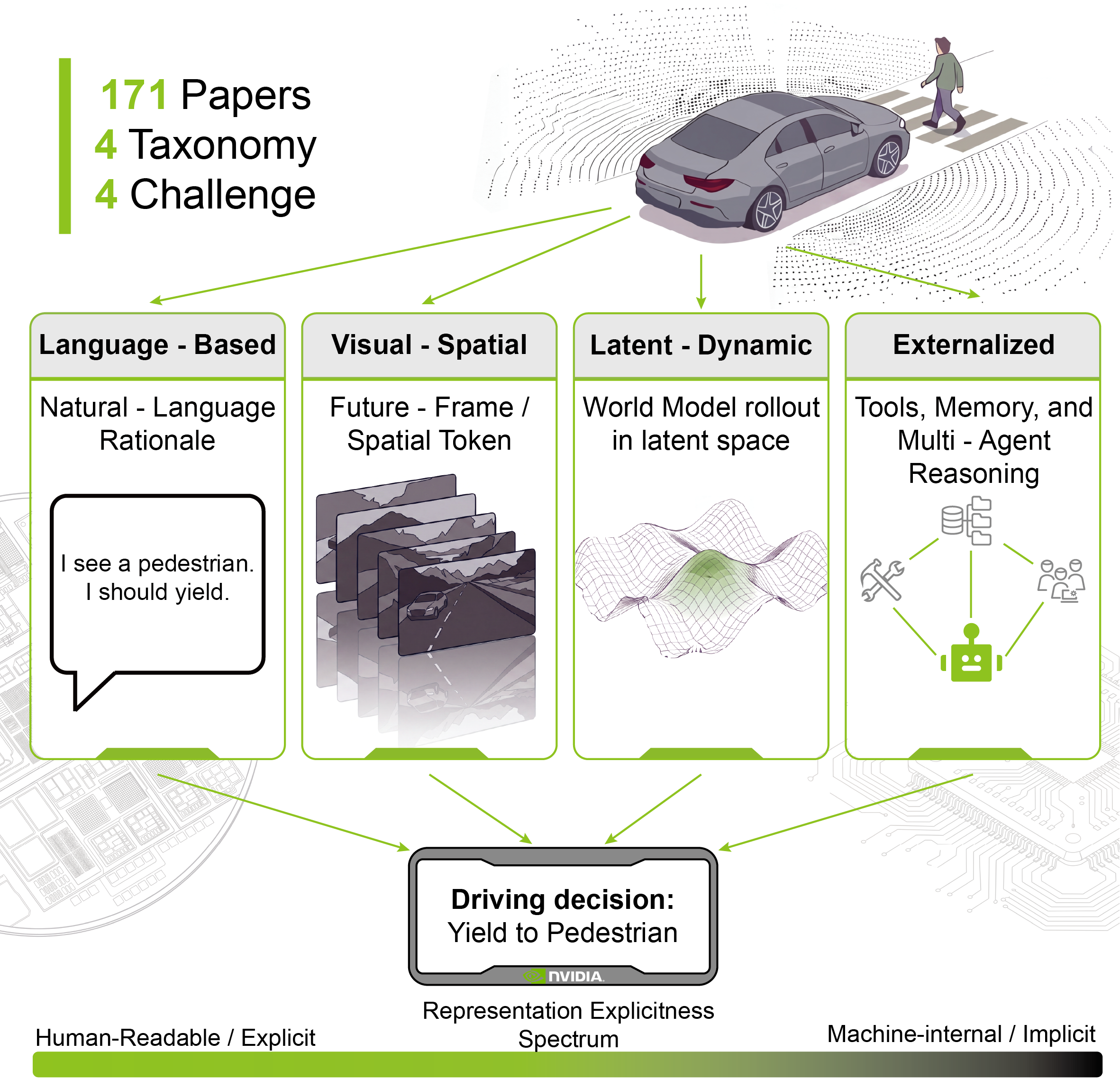}
  \caption{Overview of action-grounded reasoning representations in autonomous driving. 
Different representations can connect the same scene to a driving decision. 
They differ in interpretability, grounding, action coupling, latency, and closed-loop reliability.}
  \label{fig:carrier-overview}
\end{figure}

Chain-of-thought reasoning offers a natural starting point for such a layer. In language models, CoT encourages step-by-step reasoning before an answer~\citep{wei2022cot,wang2023selfcons,yao2023tot}, and early driving methods similarly use textual chains to describe scenes, identify critical objects, infer interactions, and derive plans~\citep{mao2023gptdriver,wang2024drivecot}. However, driving reshapes the role of CoT in three ways. First, the output is a trajectory, not a textual answer, and the two domains have different similarity structures. Second, reasoning must be grounded in geometry, motion, and traffic rules. Third, reasoning must run under real-time and closed-loop constraints, where verbose or unstable traces compound rather than reduce error.

Driving CoT is moving beyond textual chains in response to these challenges. Recent methods express intermediate reasoning as stage-wise language plans, reflective revisions, compressed textual traces, future visual states, occupancy structures, spatial crops, latent world-model rollouts, dynamics tokens, retrieved memories, traffic-rule knowledge, tool outputs, or messages exchanged among agents. Although these methods differ architecturally, they answer the same question: in what form should a driving model represent intermediate reasoning so that it can be grounded, executed, and evaluated? We call such forms action-grounded reasoning representations: identifiable intermediate structures that organize driving-relevant information before a final decision is produced.

This representation-centered view differs from existing surveys; Table~\ref{tab:survey-comparison} gives a side-by-side comparison. Our claim is not that earlier surveys omit visual, latent, externalized, or safety-related work. They provide complementary model-, task-, pipeline-, and capability-centered views~\citep{wang2024llm4ad,cotadsurvey2025,vla4adsurvey2025,liao2025e3ad,reasoningadsurvey2026}. Our survey instead asks what identifiable artifact appears between observation and a driving-relevant output, how it connects to the downstream task, and what evidence is needed to evaluate it. This common unit allows textual chains, imagined futures, latent rollouts, retrieval traces, tool calls, and multi-agent messages to be compared within a unified framework.

\begin{table*}[t]
\centering
\scriptsize
\setlength{\tabcolsep}{5pt}
\begin{tabular}{@{}p{0.27\textwidth} p{0.20\textwidth} p{0.23\textwidth} p{0.24\textwidth}@{}}
\toprule
Survey & Scope & Primary organizing axis & Treatment of intermediate representations \\
\midrule
LLM4AD~\citep{wang2024llm4ad} & LLMs for autonomous driving & LLM roles, tasks, benchmarks, and pipeline integration & Reasoning is one capability among several \\
Chain-of-Thought for Autonomous Driving~\citep{cotadsurvey2025} & Driving CoT methods & CoT paradigms, methods, and tasks & Closest in scope, but mainly organized around CoT methodology \\
\citet{vla4adsurvey2025} & VLA models for autonomous driving & Architectures and system components & Representations are discussed through VLA architecture \\
\citet{liao2025e3ad} & Past and future VLA systems & System paradigms and action generation & Focuses on model evolution and end-to-end integration \\
\citet{reasoningadsurvey2026} & Reasoning in autonomous driving & Cognitive capabilities and reasoning challenges & Organizes reasoning by capability and system challenge \\
\midrule
Ours & Decision-relevant intermediate reasoning & Language, visual-spatial, latent-dynamic, and externalized artifacts & Uses intermediate-artifact form as the paper-level axis across 13 subtypes \\
\bottomrule
\end{tabular}
\caption{Side-by-side comparison with prior surveys.}
\label{tab:survey-comparison}
\end{table*}

We include a method only when its intermediate structure plays an explicit reasoning role between observation and action: it generates or supervises an intermediate reasoning trace; it uses an intermediate state as a decision-relevant reasoning representation rather than a generic feature; or its intermediate representation can be inspected, intervened on, retrieved, compressed, distilled, refined, or communicated as part of decision-making. This boundary admits broader forms of driving CoT while excluding ordinary perception, prediction, or planning modules that do not model reasoning as an identifiable intermediate process.

\looseness=-1 Our contributions are threefold. First, we propose a representation-centered framework for driving CoT, shifting the focus from textual to action-grounded reasoning representations. Second, we organize 130 method papers into four representation families and 13 subtypes, identifying how language-based, visual-spatial, latent-dynamic, and externalized reasoning occupy different points in the trade-off space. Third, we synthesize benchmark evidence and open challenges, showing that the central obstacles for driving CoT are faithful action coupling, perceptual grounding, adaptive reasoning cost, closed-loop reliability, and safety verification.


\section{From Textual CoT to Action-Grounded Representations}
\label{sec:background}

The notion of CoT is often associated with visible natural-language reasoning~\citep{wei2022cot,wang2023selfcons,yao2023tot}. This association is useful for language tasks, where an intermediate text trace can expose how a model decomposes a question before generating an answer. In autonomous driving, however, the same definition is too narrow. A driving system may reason through language, but it may also reason through spatial evidence, future states, latent dynamics, retrieved memories, or tool outputs~\citep{zhang2023mmcot,saycan2022,innermonologue2022,rt2}. What matters is not whether the reasoning is written in natural language, but whether it forms a decision-relevant intermediate structure bridging observation and action.

\paragraph{Driving changes the output of reasoning.}

In standard language tasks, a reasoning chain connects a question to a textual answer~\citep{wei2022cot}. In driving, the output may be a trajectory, control command, maneuver decision, or planner-ready intervention~\citep{mao2023gptdriver,tian2024drivevlm,wen2024dilu}. This shifts not just the output format but the underlying domain: text and action carry different similarity structures. A trash bin and a stalled car are linguistically distant, yet operationally identical to avoid. A textual explanation that does not influence the planner may be interpretable but not action-grounded; conversely, a latent rollout that is not human-readable may still be valuable if it improves collision avoidance, comfort, and route completion~\citep{yang2025worldrft,zheng2025world4drive}. Driving CoT should be evaluated by its connection to action, not by textual coherence alone.
\vspace{-0.4em}

\paragraph{Driving changes the grounding of reasoning.}
Language CoT mainly operates over symbolic statements and commonsense knowledge~\citep{wei2022cot}. Driving reasoning must additionally be grounded in metric geometry, lane topology, map priors, object locations, occlusion, motion, interaction intent, and traffic rules~\citep{tian2025nuscenesspatialqa,godbole2025dramax}. A plausible chain can be unsafe if it mislocates a pedestrian, ignores a blind spot, or misunderstands right-of-way; this grounding pressure explains why many methods move beyond text. Some methods attach language reasoning to visual regions, object crops, masks, or bounding boxes~\citep{corbiere2025retrievalbased,li2025spacedrive}. Others imagine future visual states or occupancy structures~\citep{wang2023drivewm,zeng2025futuresightdrive,yang2024driveoccworld}. Still others compress scene evolution into latent world models or dynamics tokens~\citep{zheng2025world4drive,shang2026dynvla,team2026onevl}. These designs are not alternative explanation formats but attempts to move reasoning closer to the variables that determine safe driving.

\paragraph{Driving changes the deployment constraints of reasoning.}
Language reasoning can be long and reflective in offline tasks, but driving operates under real-time and closed-loop constraints~\citep{zhang2025bench2advlm,jia2024bench2drive}: a slow, detailed trace may help debugging yet harm execution, and the vehicle's action changes the next observation, so reasoning errors compound through interaction. This motivates compressed, distilled, latent, and adaptive forms of CoT. Some systems use explicit textual reasoning as supervision during training and internalize it into faster representations at inference time~\citep{feng2025verdi,liu2025dsdrive}. Others trigger expensive reasoning only in uncertain or high-risk scenarios~\citep{qian2024fasionad,luo2025adathinkdrive}. The central question is not whether a model can reason more, but whether it reasons at the right level of detail under the right conditions.

\section{Representation-Centered Taxonomy}
\label{sec:taxonomy}

\vspace{-0.5em}

Figure~\ref{fig:representation-trend} (Appendix~\ref{app:comparison-table}) gives a corpus-level view: growth accelerates after 2025, and while language-based methods dominate, externalized, latent-dynamic, and visual-spatial representations now form substantial branches. This pattern motivates our four-way organization into language-based~\citep{mao2023gptdriver,tian2024drivevlm,yuan2025autodriver,feng2025verdi}, visual-spatial~\citep{wang2023drivewm,yang2024driveoccworld,corbiere2025retrievalbased,li2025spacedrive}, latent-dynamic~\citep{zheng2025world4drive,shang2026dynvla,yang2025worldrft}, and externalized~\citep{wen2024dilu,luo2025senserag,qian2025agentthink,chiu2025v2vllm} representations. These categories are not a strict temporal progression; they reflect different design priorities: interpretability, grounding, action coupling, efficiency, knowledge access, and collaboration.

\vspace{-0.6em}

We assign the primary label according to the form in which the decision-relevant intermediate artifact is consumed at the reasoning-to-output interface, rather than its surface serialization or provenance. Tool provenance is noted separately where relevant, but does not by itself determine the primary family. For example, a textual coordinate description that subsequent reasoning consumes as tokens is language-based despite its spatial content; the same values parsed into a bounding box, object-state tuple, occupancy element, or waypoint consumed by prediction or planning are visual-spatial, even if serialized as text; and a tool-returned bounding box consumed as structured spatial state likewise remains visual-spatial, with its tool provenance noted separately. This section discusses each subtype at the level of common design patterns and representative trade-offs; Table~\ref{tab:representative} gives one representative method per subtype, and full paper-level comparisons, including backbone, benchmark, and assigned subtype, are provided in Appendix~\ref{app:comparison-table}.

\begin{table*}[t]
\centering
\scriptsize
\setlength{\tabcolsep}{3pt}
\renewcommand{\arraystretch}{1.0}
\begin{tabular}{@{}>{\raggedright\arraybackslash}p{0.10\textwidth} >{\raggedright\arraybackslash}p{0.115\textwidth} >{\raggedright\arraybackslash}p{0.225\textwidth} >{\raggedright\arraybackslash}p{0.10\textwidth} >{\raggedright\arraybackslash}p{0.13\textwidth} >{\raggedright\arraybackslash}p{0.26\textwidth}@{}}
\toprule
Subtype & Representative & Intermediate artifact & Backbone & Benchmark/task & Source-reported within-study evidence \\
\midrule
\multicolumn{6}{@{}l}{\textit{Language-based}} \\
Descriptive & ORION~\citep{fu2025orion} & Scene description and causal rationale aligned with trajectory head & Vicuna & Bench2Drive closed-loop & Driving Score 77.74; Success Rate 54.62\% \\
Procedural & DriveVLM~\citep{tian2024drivevlm} & Stage-wise perception--prediction--planning language plan & Qwen & nuScenes open-loop planning & Average L2 0.40\,m; 0.31\,m for the dual-system variant in the same study \\
Reflective & AutoDrive-R\textsuperscript{2}~\citep{yuan2025autodriver} & Critique followed by a revised reasoning trace & Qwen & nuScenes open-loop planning & Average L2 0.19\,m; matched base model 1.45\,m \\
Compressed & AdaThinkDrive~\citep{luo2025adathinkdrive} & Adaptive think/no-think textual trace & InternVL & NAVSIM v1 navtest & PDMS 90.3 with adaptive CoT switching \\
\midrule
\multicolumn{6}{@{}l}{\textit{Visual-spatial}} \\
Predictive & FSDrive~\citep{zeng2025futuresightdrive} & Predicted future-scene representation & Qwen & nuScenes open-loop planning & Average L2 0.28\,m; collision rate 0.10\% \\
Action-grounded & RIV-CoT~\citep{corbiere2025retrievalbased} & Retrieved and cropped visual evidence interleaved with the textual chain & Qwen+LLaVA & DrivingVQA & F1 68.8 with interleaved visual CoT \\
\midrule
\multicolumn{6}{@{}l}{\textit{Latent-dynamic}} \\
Rollout-based & World4Drive~\citep{zheng2025world4drive} & Intention-conditioned latent world-model rollouts & E2E CNN/Trans. & NAVSIM v1 navtest & PDMS 85.1 without perception annotation \\
Tokenized & OneVL~\citep{team2026onevl} & Four visual and two language latent positions & Qwen & NAVSIM planning & PDM score 88.84 at 4.46\,s; explicit-CoT baseline 88.29 at 6.58\,s \\
Optimization-based & ReCogDrive~\citep{li2025recogdrive} & Diffusion-refined trajectory states over cognitive features & Qwen+InternVL & NAVSIM v1 navtest & PDMS 90.8 \\
\midrule
\multicolumn{6}{@{}l}{\textit{Externalized}} \\
Retrieval-augmented & DiLu~\citep{wen2024dilu} & Retrieved past driving experiences injected into the prompt & GPT & HighwayEnv closed-loop & Success rate improves as the experience memory accumulates \\
Knowledge-grounded & LeAD~\citep{zhang2025lead} & Traffic-law and map priors guiding hierarchical planning & GPT & CARLA Leaderboard V1 & Driving Score 71.96 (closed-loop) \\
Tool-mediated & AgentThink~\citep{qian2025agentthink} & Typed tool calls and returned observations & Qwen & DriveLMM-o1 reasoning & Reasoning score 79.68; untuned backbone 51.77 \\
Cooperative & CoLMDriver~\citep{liu2025colmdriver} & Inter-vehicle natural-language negotiation messages & InternVL & CARLA InterDrive closed-loop & Driving Score 88.53 across interactive scenarios \\
\bottomrule
\end{tabular}
\caption{One representative method per subtype. Representatives are selected for the clarity of their intermediate artifact and the availability of a documented within-study comparison, not for the highest reported score. Each result retains the source paper's model configuration, split, horizon, inputs, and metric definition; rows are not ranked or compared across incompatible protocols, and baselines are included only when reported within the same study or under a matched setting. Appendix~\ref{app:comparison-table} remains the complete corpus inventory.}
\label{tab:representative}
\end{table*}

\subsection{Language-Based Representations}
\label{sec:language}

Language-based representations are the most direct extension of standard CoT to driving~\citep{wei2022cot,mao2023gptdriver,xu2023drivegpt4}, expressing intermediate reasoning as scene descriptions, object analyses, risk explanations, decision justifications, driving plans, self-reflections, or compressed textual traces~\citep{tian2024drivevlm,wang2024drivecot,yuan2025autodriver,jiang2025alphadrive}. Their main advantage is inspectability: language traces can be read, aligned with traffic rules, supervised by annotations, and used for passenger-facing explanations~\citep{hwang2024emma,wen2024dilu,cui2023receive}. Their main weakness is action coupling: language describes geometry and motion indirectly~\citep{tian2025nuscenesspatialqa,drivemind2025}.

\paragraph{Descriptive language reasoning.}
Descriptive language reasoning verbalizes scene semantics, critical objects, risk factors, and driving decisions before producing an action or explanation~\citep{mao2023gptdriver,xu2023drivegpt4,sha2023languagempc,ma2023dolphins,hwang2024emma}. This subtype makes driving models more transparent by exposing what the model appears to notice and why it chooses a maneuver~\citep{zheng2024simplellm4ad,yao2024calmmdrive,zhang2024wisead,xing2024openemma,fu2025orion,chahe2025reasondrive,liu2025xdriver,lu2025realad}. It is especially useful for driving question answering, explanation, and instruction-following~\citep{chi2025impromptu,zarghani2025multimodal,liu2025omnireason,peng2025system,yu2025robust,wu2025enhancing,zhang2025cocvla,liao2025e3ad,guo2026listen,gao2026steervlab,xie2025deliberation}. However, description alone does not guarantee action coupling: a model may describe a scene correctly yet output a poor trajectory. The core risk is post-hoc rationalization: a text trace may look like reasoning even when the action was produced by a separate planner or action head.

\paragraph{Procedural language reasoning.}
Procedural language reasoning imposes an explicit reasoning structure, often following perception, prediction, planning, and control stages~\citep{wang2023drivemlm,tian2024drivevlm,wang2024drivecot,luo2024pkrdcot,mandalika2025primedrivecot}. Compared with descriptive reasoning, procedural reasoning is more action-aware because it organizes information in a decision-relevant order and can be connected to planners or behavior states~\citep{sharan2024llmassist,peng2024lcllm,wang2024dualad,zhao2025sce2drivex,liao2025cotdrive,diao2025driverx}. Its limitation is rigidity: fixed templates may over-reason in simple scenarios and under-specify complex negotiation, while perception or grounding errors can still propagate through a well-structured but incorrect chain. A promising direction is adaptive procedure selection across lane following, merging, occluded hazards, and emergencies~\citep{wang2025hmvlm,liu2025reasonplan,wang2025cot4ad,liu2025llavida,han2026applevlm,tao2026navidrivevlm,ghosh2026radlad,gu2026accelerating}. Hybrid procedural reasoning can combine symbolic calculation with commonsense guidance. \citet{azarafza2024hybrid} provide an LLM with detected objects and vehicle-state signals, then use an explicit sequence of arithmetic and commonsense reasoning steps to produce brake and speed commands in CARLA. We assign procedural-language reasoning as the primary subtype because this staged procedure is the intermediate artifact that structures control generation, and record knowledge-grounded reasoning as a secondary tag because the procedure also invokes commonsense driving knowledge.

\paragraph{Reflective language reasoning.}
Reflective language reasoning adds critique, revision, rollback, or failure learning to the textual reasoning process~\citep{yuan2025autodriver,li2025reflectdrive,zhang2025openread}. This turns CoT from a one-pass explanation into an error-correction mechanism, valuable in interactive or high-risk scenarios where an initial plan may be inconsistent, unsafe, or incomplete~\citep{peng2025counterfactual,luo2026unleashing}. The central trade-off is latency: reflection requires additional generation, verification, or optimization steps, making it difficult to deploy at every time step. A practical reflective system must decide when deeper reasoning is necessary and when fast execution is safer; reflection is thus best viewed as a risk-triggered mechanism for difficult scenes rather than a universal inference mode.

\paragraph{Compressed language reasoning.}
Compressed language reasoning shortens, distills, internalizes, or selectively activates textual reasoning to reduce inference cost~\citep{huang2024making,jiang2025alphadrive,qiao2025lightemma,liu2025dsdrive,feng2025verdi,wasif2025drivemind}. These methods reveal that textual CoT may be most useful not as a verbose inference-time output, but as a training-time scaffold~\citep{zhou2025autovla,li2025driver1,zheng2025driveagentr1,luo2025adathinkdrive,nvidia2025alpamayor1}. A model can learn from explicit reasoning traces and later execute behavior through shorter language, latent states, or direct action heads~\citep{zhang2025reasoningvla,li2025covlmrl,fu2025minddrive,zhang2025omnidriver1,zhao2026thinkdrive,chen2026bridging,ye2026autodrivep3}. This blurs the boundary with latent-dynamic reasoning and reframes the goal: not longer chains, but better transfer from linguistic reasoning to action-effective representations that preserves the benefits of explicit supervision after reasoning is internalized.

\subsection{Visual-Spatial Representations}
\label{sec:visual}

Visual-spatial representations move intermediate reasoning closer to the perceptual structure of driving scenes, treating future frames, occupancy maps, object crops, masks, bounding boxes, or trajectory sketches as reasoning carriers. This addresses a key weakness of text: language describes spatial relations, but does not preserve metric geometry, occlusion, lane topology, or temporal evolution.

\paragraph{Predictive visual-spatial reasoning.}
Predictive visual-spatial reasoning represents intermediate reasoning as future visual or occupancy states~\citep{wang2023drivewm,gao2024vista,yang2024driveoccworld,chen2024drivinggpt}. These methods allow the model to see before acting by simulating how the scene may evolve under possible ego actions~\citep{zeng2025futuresightdrive,xiong2026unidrivewm,zhou2026drivedreamerpolicy,wang2026vlaworld}, which is attractive for interaction-heavy scenarios such as merging, unprotected turns, and pedestrian crossings. The limitation is that visual plausibility does not imply planning usefulness: a realistic future may fail to preserve the geometry, uncertainty, or agent behavior needed for safe control, so the key question is whether predicted futures causally improve driving decisions rather than video or occupancy metrics.

\paragraph{Action-grounded visual-spatial reasoning.}
\looseness=-1 Action-grounded visual-spatial reasoning localizes the evidence most relevant to the current action, such as critical-object crops, boxes, masks, depth-aware spatial features, or trajectory sketches~\citep{corbiere2025retrievalbased,chen2025solve,li2025spacedrive}. Compared with full future generation, this subtype is more compact and more directly tied to the final maneuver, focusing the model on the pedestrian to avoid, the vehicle to yield to, or the trajectory region to refine~\citep{min2026visionlanguageaction,zhang2026minddriver}. Its risk is evidence selection: if the crop, box, mask, or sketch misses the true risk factor, later reasoning may become confidently wrong. This shifts the burden from complete simulation to faithful selection of decision-relevant evidence; for safety-critical deployment, the system must also know when its selected evidence is incomplete.

\subsection{Latent-Dynamic Representations}
\label{sec:latent}

Latent-dynamic representations compress reasoning into latent states, dynamics tokens, rollouts, or iterative optimization processes. They are less interpretable than text or images, but often more compatible with planning and control: reasoning need not be human-readable to be decision-relevant.

\paragraph{Rollout-based latent-dynamic reasoning.}
Rollout-based latent-dynamic reasoning uses latent world models to simulate future scene evolution before action selection~\citep{bytedance2025uniugp,zheng2025world4drive,lin2025futurex,latentcotwm2025}. This preserves thinking through possible futures without pixel-level generation cost: latent rollouts represent temporal dynamics more compactly than videos and more action-relevantly than language~\citep{liao2025think,li2026sgdrive,jia2026driveworldvla,luo2026lastvla}. The main challenge is verification: a human can inspect a textual chain or predicted frame, but not a latent rollout. Future systems need probing, decoding, consistency checks, or intervention tests to determine whether latent reasoning is faithful, grounded, and safe.

\paragraph{Tokenized latent-dynamic reasoning.}
Tokenized latent-dynamic reasoning uses compact discrete codes or continuous latent positions explicitly trained to carry information about future dynamics, actions, or intermediate reasoning~\citep{ganai2026rbencore,wang2026histvla,shang2026dynvla,team2026onevl}. DynVLA quantizes future ego and environment dynamics into codes supervised through future-scene reconstruction and action prediction~\citep{shang2026dynvla}, while OneVL uses a small fixed set of visual and language latent positions trained through reconstruction objectives and consumed before action generation~\citep{team2026onevl}. Unlike generic hidden features, these carriers are identifiable in the architecture and are explicitly supervised, decoded, or passed to a downstream module. We also clarify that HiST-VLA is a hybrid boundary case: it produces an explicit granular command, a coarse trajectory, and one aggregate trajectory-confidence score, rather than attaching a confidence value to each primitive or representing the explicit command labels as learned dynamics tokens~\citep{wang2026histvla}. This subtype inherits the sequential structure of textual CoT while removing the requirement that each token be a word, but its open problem is semantic alignment: without probing or intervention, it is unclear whether latent carriers encode risk, intention, affordance, motion, or statistical shortcuts.

\paragraph{Optimization-based latent-dynamic reasoning.}
Optimization-based latent-dynamic reasoning treats reasoning as iterative refinement of latent action states, often through diffusion, reinforcement fine-tuning, parallel decoding, or progressive trajectory optimization~\citep{jiang2025diffvla,gao2025diffvla,peng2025colavla}. In this view, the reasoning trace is not a sentence or image, but a trajectory of internal optimization states. This is attractive because planning is naturally an optimization problem: a candidate trajectory can be refined to better satisfy safety, comfort, and goal constraints~\citep{li2025recogdrive,ma2025dvlmad,yang2025worldrft}. Its limitation is observability: if refinement improves scores but cannot be inspected or causally linked to risk reduction, it is hard to certify as reasoning rather than opaque optimization---an important boundary for future work.

\subsection{Externalized Representations}
\label{sec:external}

Externalized representations move part of reasoning outside the model's internal activations, using retrieval, memory, traffic rules, tools, or communication among agents instead of parametric knowledge alone. Their strength is long-tail coverage and modularity; their weakness is reliability, because external sources may be stale, irrelevant, delayed, inconsistent, or unsafe.

\paragraph{Retrieval-augmented reasoning.}
Retrieval-augmented reasoning uses retrieved regulations, demonstrations, cases, experiences, or vehicle-to-everything information as intermediate evidence for driving decisions~\citep{cai2024regulation,luo2025senserag,han2025traffic,wang2025rad}. Retrieval helps with long-tail scenarios where parametric memory is insufficient, such as rare traffic rules, unusual road layouts, or previously seen risky interactions~\citep{chang2025drivingrag,luo2025v2xunipool,gan2025casebased,patrikar2025case}, and makes part of reasoning auditable because retrieved evidence can be inspected. The new failure mode is retrieving irrelevant, outdated, or misleading evidence and reasoning confidently from it, so retrieval quality, source reliability, and fallback behavior are central to this subtype.

\paragraph{Knowledge-grounded reasoning.}
Knowledge-grounded reasoning uses explicit rules, maps, scene graphs, risk constraints, or other structured priors that cannot be inferred reliably from the current observation alone~\citep{cui2023receive,jiang2024koma,xu2025telldrive,fang2025interact,xu2025humancentric}. SafeDrive turns risk knowledge into decision constraints~\citep{zhou2024safedrive}; PlanAgent and LeAD expose route or map priors to planning~\citep{zheng2024planagent,zhang2025lead}; and KLDrive constructs a scene knowledge graph for constrained reasoning~\citep{tian2026kldrive}. We distinguish this subtype from retrieval by the role of the intermediate artifact: retrieval is primary when memory access and example selection constitute the reasoning process, whereas knowledge-grounded reasoning is primary when a rule, graph, or structured prior itself guides the decision. Memory-centric methods such as DiLu and Agent-Driver are therefore retrieval-primary, with a secondary knowledge tag where appropriate~\citep{wen2024dilu,mao2024agent}. Mechanisms reported by individual papers, such as confidence-weighted guidance or lower-level overrides, are described as system-specific safeguards rather than evidence of a general solution to knowledge--scene conflict~\citep{qian2024fasionad,qian2025fasionad,liu2025vlmudmc,luo2025mtrdrive,wang2025kept,zhang2025adadrive}.

\paragraph{Tool-mediated reasoning.}
Tool-mediated reasoning delegates part of the reasoning process to callable modules such as planners, simulators, map queries, rule checkers, or structured APIs~\citep{qian2025agentthink,goba2026prompts}, making driving CoT more executable: instead of only stating a plan, the model can call a tool that computes, checks, or simulates part of the decision. Tool use also offers modularity, since specialized components handle geometry, rules, or optimization more reliably than a general-purpose model. The cost is interface reliability: a call may fail, return delayed output, or be invoked in the wrong context, so tool-mediated CoT requires validation, uncertainty handling, and safe fallback policies.

\paragraph{Cooperative reasoning.}
\looseness=-1 Cooperative reasoning represents intermediate reasoning as messages exchanged among vehicles, agents, or infrastructure components~\citep{hu2024agentscodriver,chiu2025v2vllm}. Many driving risks are distributed: no single vehicle observes the whole scene, and cooperative negotiation can improve merging, intersection handling, or occluded-risk awareness. Communication exposes intent and shares local observations, making reasoning more socially and spatially informed~\citep{liu2025colmdriver,gao2025langcoop,hou2025driveagent}. However, cooperation introduces synchronization, authentication, bandwidth, and conflict-resolution challenges: the system must decide which messages to trust and how to remain safe when communication is missing or adversarial.

\section{Cross-Category Analysis}
\label{sec:analysis}

The four categories are not competing replacements; they answer the same pressure: intermediate reasoning must be both meaningful to humans and useful for action. Language-based reasoning is easiest to inspect but weakest in metric grounding~\citep{mao2023gptdriver,tian2024drivevlm,hwang2024emma}; visual-spatial reasoning is more grounded but can be costly or incomplete~\citep{corbiere2025retrievalbased,li2025spacedrive,zeng2025futuresightdrive}; latent-dynamic reasoning is closer to control but harder to verify~\citep{zheng2025world4drive,yang2025worldrft,luo2026lastvla}; and externalized reasoning expands access to knowledge and cooperation but depends on external sources and interfaces~\citep{wen2024dilu,luo2025senserag,qian2025agentthink}.

\paragraph{Interpretability versus action coupling.}
A central tension is that the most interpretable representations are not always the most action-effective. Textual chains expose the model's apparent logic but may remain loosely connected to the final trajectory~\citep{drivemind2025,tian2024drivevlm,mao2023gptdriver}; latent rollouts and optimization states can directly influence planning but are difficult to inspect~\citep{yang2025worldrft,li2025recogdrive,gao2025diffvla}. Visual and externalized representations occupy a middle ground: crops, masks, retrieved cases, and tool outputs are easier to inspect than latent states, but still require mechanisms to verify that the final action actually depends on them~\citep{corbiere2025retrievalbased,min2026visionlanguageaction,chang2025drivingrag,patrikar2025case,qian2025agentthink}. Future systems may therefore need dual representations: a compact internal representation for control and a faithful external trace for monitoring and explanation.

\paragraph{Grounding versus coverage.}
Another tension is between grounding and coverage. Visual-spatial reasoning grounds decisions in the current scene but may miss long-tail knowledge such as rare rules or unusual interactions~\citep{corbiere2025retrievalbased,li2025spacedrive,tian2025nuscenesspatialqa,godbole2025dramax}; retrieval and knowledge-grounded reasoning provide this coverage but may introduce stale or irrelevant information~\citep{cai2024regulation,luo2025senserag,chang2025drivingrag,wen2024dilu}; latent world models encode temporal dynamics but may fail silently under distribution shift~\citep{wang2023drivewm,gao2024vista}. A robust driving CoT system should integrate perception-grounded evidence, learned dynamics, and external knowledge.

\paragraph{Reasoning depth versus latency.}
Reasoning is not free: reflection, retrieval, tool use, world-model rollout, and multi-agent communication all increase computation or communication cost~\citep{yuan2025autodriver,li2025reflectdrive,luo2025senserag,qian2025agentthink,gao2025langcoop}. A system that reasons too slowly may fail in urgent situations, while one that reasons too little may miss rare hazards. This points to adaptive reasoning as a unifying direction~\citep{luo2025adathinkdrive,qian2024fasionad,xie2025deliberation,ghosh2026radlad}: estimate uncertainty, risk, and conflict, then decide whether to use fast reactive control, short language reasoning, retrieved knowledge, latent simulation, tool-mediated verification, or cooperative communication.

\paragraph{From representation choice to system design.}
Representation choice is a system-design decision rather than a purely modeling one; four questions make it concrete. \textbf{Trigger}: When is additional reasoning activated---for example, under uncertainty, ambiguous right of way, or conflict between scene evidence and a retrieved rule~\citep{luo2025adathinkdrive,qian2024fasionad}? \textbf{Interface}: Where does the intermediate artifact affect the task---for example, maneuver selection, trajectory scoring, constraint construction, or candidate rollout---rather than appearing only as a post-hoc explanation? \textbf{Validation}: What is checked before the artifact is used, such as entity grounding for language traces, geometric and temporal consistency for spatial states, rollout consistency for latent states, or source and scene compatibility for retrieved information? \textbf{Fallback}: What alternative system behavior is used when the artifact is missing, inconsistent, uncertain, or late, and under what assumptions is that alternative evaluated? Inconsistencies among perception, the intermediate artifact, and the planned action should be checked separately; the planned action is not treated as ground truth. This checklist is design and evaluation guidance, not evidence that the surveyed systems already satisfy deployment-level safety requirements.

\section{Evaluation Landscape}
\label{sec:benchmarks}

Existing benchmarks evaluate different fragments of driving reasoning. QA and explicit-reasoning datasets test whether models can describe scenes, answer questions, and produce structured reasoning traces~\citep{sima2023drivelm,wang2024omnidrive,wei2025driveqa,ishaq2025drivelmmo1,zhang2025bench2advlm,zeng2025dvbench}. Spatial grounding and robustness benchmarks test whether models localize objects, understand relations, and remain stable under perturbations~\citep{tian2025nuscenesspatialqa,alvar2025segments,li2025finegrained}. Planning benchmarks evaluate whether reasoning improves motion decisions, collision avoidance, route, and progress~\citep{navsim,jia2024bench2drive}, and closed-loop benchmarks expose whether reasoning remains reliable when actions affect future observations~\citep{inoue2023evaluation,wei2025adbench}.

Despite this progress, evaluation remains fragmented: QA benchmarks rarely test action consequences, planning benchmarks rarely inspect intermediate reasoning, and closed-loop benchmarks rarely evaluate reasoning faithfulness or evidence reliability (details in Appendix~\ref{app:benchmark-details}). Evaluation should jointly measure grounding, action faithfulness, reasoning cost, and closed-loop safety.

QA and planning metrics measure different parts of a reasoning system. Exact-match or multiple-choice accuracy measures final-answer correctness, but does not show whether the answer is visually grounded or used by the planner. Trajectory displacement metrics compare a prediction with a logged future and may penalize other safe behaviors. Open-loop collision estimates depend on the footprint, horizon, object extrapolation, and collision protocol, and do not model how other agents respond to the ego vehicle. Closed-loop and composite scores cover more aspects of driving, but remain specific to the simulator, benchmark version, scenario set, and aggregation rule. We therefore report these metrics as source-specific evidence rather than treating them as directly comparable measures of reasoning quality. A compact metric glossary is provided in Appendix~\ref{app:metric-glossary}.

\section{Open Challenges}
\label{sec:challenges}

\paragraph{Faithful action coupling.}
\looseness=-1 The most important question for driving CoT is not whether the reasoning trace is plausible, but whether it actually affects action: a model may generate a convincing explanation the planner ignores, or act safely for reasons that differ from the stated chain. Future work should evaluate causal coupling through intervention: changing the reasoning representation should predictably change the action, and changing irrelevant text should not. This is especially important for language-based explanations, where post-hoc rationalization can be mistaken for faithful reasoning. Two safeguards make such intervention tests meaningful. First, the intervention must preserve artifact validity rather than create an implausible or out-of-distribution state. Second, the expected behavioral change is assessed at the artifact's abstraction level, not necessarily by exact waypoint correspondence: high-level reasoning may affect maneuver class, yielding behavior, risk profile, trajectory family, control decision, or closed-loop outcome through a downstream planner, vehicle dynamics, map constraints, or a safety override.

\paragraph{Adaptive reasoning budget.}
Routine lane following may require fast reactive control, while ambiguous intersections or rare hazards may require deeper deliberation, retrieval, or simulation. A major challenge is to allocate reasoning budget adaptively: reason more when uncertainty, risk, or conflict is high, and less when the action is obvious, which requires uncertainty estimation, risk-aware triggering, early exiting, and safe fallback.

\paragraph{Closed-loop safety evaluation.}
Open-loop benchmarks are insufficient because errors compound through interaction: a trace that improves offline question-answering may still destabilize closed-loop driving if it delays action, overreacts to spurious risks, or produces inconsistent plans across time. Closed-loop evaluation should measure not only collision and route completion, but also temporal stability of reasoning, recovery from incorrect intermediate states, and robustness under distribution shift, with monitors for abnormal reasoning such as missing risk factors, impossible future states, irrelevant retrievals, failed tool calls, or conflicting agent messages.

\subsection{Representation-Specific Verification}
\label{sec:verification}
Three properties of a reasoning artifact should be kept distinct: \emph{grounding}, whether the artifact correctly reflects the scene, its geometry, rules, and dynamics; \emph{action-faithfulness}, whether the downstream action actually relies on it; and \emph{closed-loop safety}, whether the system remains safe under interaction, latency, and error accumulation. A grounded artifact may be ignored by the planner, while an erroneous artifact may be relied upon and produce an unsafe action, so verification mechanisms must be representation-specific rather than limited to final-action metrics.
For language-based artifacts, check whether referenced objects, relations, maps, and rules agree with the scene, and whether valid changes to action-relevant content are accompanied by compatible downstream changes at the corresponding abstraction level. For visual-spatial artifacts, check geometric validity, lane topology, temporal consistency, and whether decision-critical spatial evidence is covered. For latent-dynamic artifacts, decode or probe motion and risk variables, test transition consistency, and report uncertainty or rollout disagreement. For externalized artifacts, check source provenance, freshness, context match, tool-output validity, latency, and conflicts between external information and local observations.
Inconsistencies among perception, the intermediate artifact, and the planned action are flagged for separate checking rather than assuming that any one component is ground truth. These are evaluation targets, not evidence that current systems provide formal safety guarantees.

\section{Conclusion}
\label{sec:conclusion}

Driving is action-oriented, perception-based, and latency-constrained, so intermediate reasoning has expanded beyond text into language-based, visual-spatial, latent-dynamic, and externalized representations: language supports inspection, visual evidence improves grounding, latent dynamics strengthen action coupling, and externalized reasoning unlocks knowledge and cooperation. No single representation suffices; future systems should combine them to be grounded, causally linked to behavior, efficient, and monitored in closed-loop driving. The future of driving CoT is not longer textual chains, but reasoning representations that can be executed and trusted in safety-critical systems.

\section*{Limitations}

This survey has several limitations. First, the field of CoT and foundation models for autonomous driving is evolving rapidly, and new arXiv papers and technical reports continue to appear at a high frequency. Our corpus therefore reflects the publicly available literature at the time of writing and may not cover every very recent work.
Second, our taxonomy assigns each method to its primary reasoning representation. This improves clarity, but inevitably simplifies hybrid systems that combine language reasoning, visual grounding, latent planning, retrieval, tool use, or agent communication. Alternative categorizations may be reasonable for methods whose reasoning-action pipeline relies on multiple representations.
Finally, our scope is centered on intermediate reasoning representations rather than all foundation-model-based autonomous driving research. We include perception, prediction, planning, simulation, VLA training, and world-modeling works only when their intermediate states explicitly serve a decision-relevant reasoning role. This boundary enables a focused analysis of the transition from textual CoT to action-grounded reasoning, but it also excludes some important autonomous driving methods outside this focus. This survey conducts no author-run ablation, intervention study, or matched-condition comparison and therefore makes no causal claim about the superiority or action-faithfulness of any representation family. Its original contribution is a corpus-level analysis of representation-action interfaces and of the evidence used to evaluate them.

\bibliography{corpus}

\appendix

\clearpage

\section{Four Shifts in Driving CoT}
\label{app:four-shifts}

The main body of this survey organizes driving CoT methods by the form of their intermediate reasoning representation. In this appendix, we step back from individual categories and examine the corpus from a broader historical perspective. Based on the 130 method papers surveyed in this work, we identify four shifts that jointly characterize the evolution of action-grounded reasoning in autonomous driving: an infrastructure shift from closed APIs to open backbones, a cognitive shift in what models reason about, an action shift from verbal explanation to trajectory generation, and a deployment shift from always-on reasoning to cost-aware adaptive reasoning. These shifts are not independent. Open backbones enable fine-tuning and reinforcement learning; stronger training pipelines make action-oriented reasoning possible; action-oriented reasoning exposes the faithfulness gap between explanation and control; and deployment constraints force the field to reconsider when, how, and how much a driving model should reason.


\subsection{The Infrastructure Shift: From Closed APIs to Open Backbones}
\label{app:shift-infrastructure}

The earliest wave of driving CoT methods was largely built on general-purpose large language models. Systems such as GPT-Driver and LanguageMPC serialized driving scenes into textual prompts and queried GPT-style models to obtain decisions, waypoints, or control-related outputs~\citep{mao2023gptdriver,sha2023languagempc}. This design was natural at the time: closed APIs provided strong language reasoning ability, required little training infrastructure, and allowed researchers to quickly test whether language-style reasoning could be useful for driving. However, this approach also imposed a clear ceiling. Closed APIs restrict access to model weights, limit gradient-based adaptation, make latency difficult to control, and prevent tight integration with perception and planning modules.

\begin{figure*}[t]
  \centering
  \includegraphics[width=\textwidth]{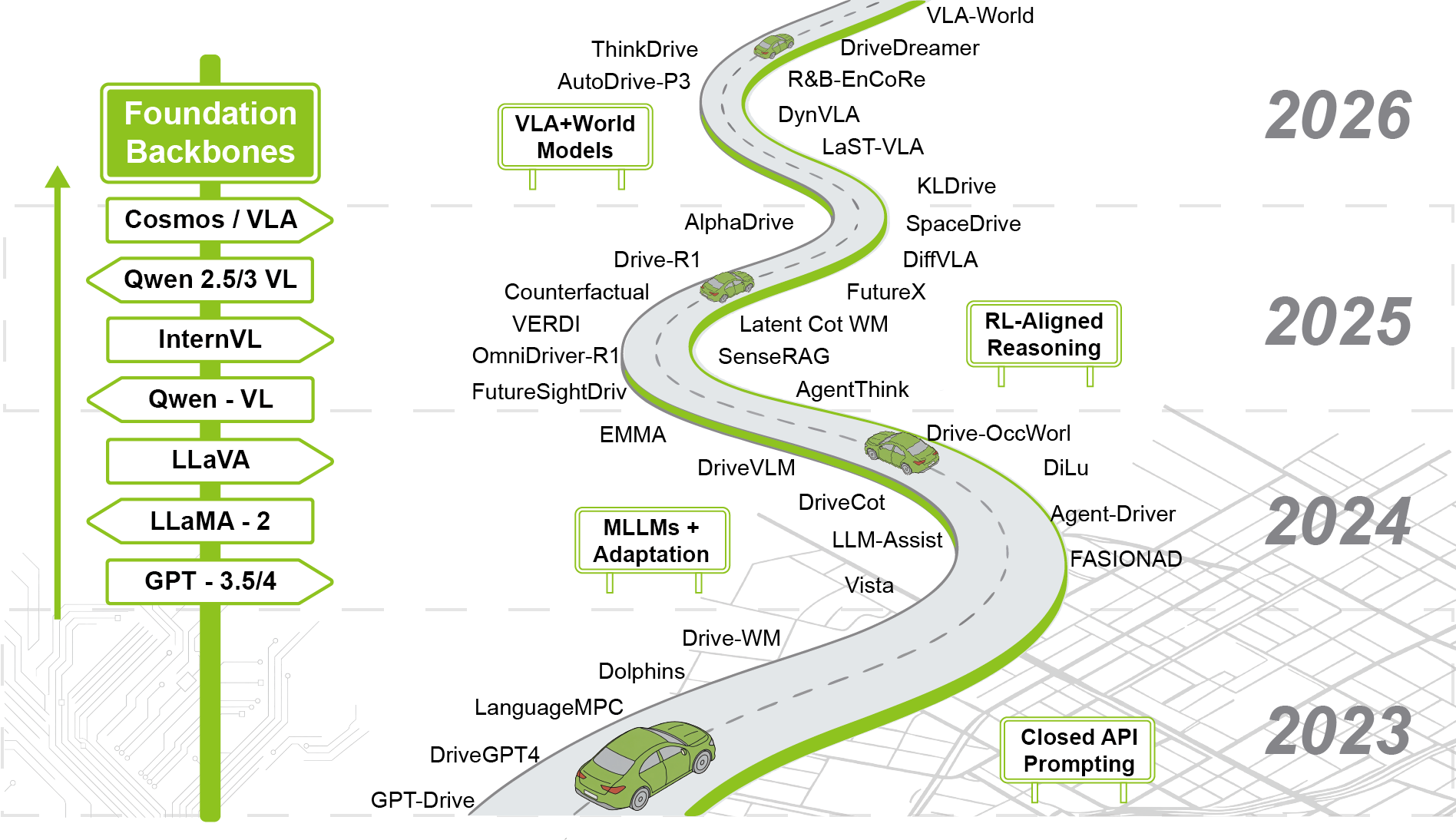}
  \caption{The infrastructure shift in driving CoT. Representative methods are placed along a timeline spanning three phases: closed-API LLMs (2023), open-weights multimodal models (2024--2025), and vision-language-action models and world models (2025--2026). Foundation models are listed on the left; the transition from prompting to SFT to RL alignment tracks directly with the shift from closed to open backbones.}
  \label{fig:shift-infrastructure}
\end{figure*}

The corpus shows a clear transition away from this paradigm. In 2023, a substantial fraction of methods still relied on GPT-3.5, GPT-4, or other closed models. By 2025 and 2026, however, the dominant backbones are open or semi-open vision-language models such as LLaVA, Qwen-VL, InternVL, and their driving-specialized variants~\citep{tian2024drivevlm,zheng2024simplellm4ad,xing2024openemma,liu2025llavida}. This change is not merely a preference for open-source implementation. It changes the kind of research that becomes possible. Once model weights are available, researchers can perform supervised fine-tuning on driving-specific reasoning traces, distill reasoning into smaller models, align models with trajectory-level objectives, or optimize reasoning behavior with reinforcement learning~\citep{feng2025verdi,liu2025dsdrive,li2025driver1,zheng2025driveagentr1,zhang2025omnidriver1}.

The rise of reinforcement learning further strengthens this trend. Methods such as Drive-R1, DriveAgent-R1, OmniDrive-R1, MindDrive, and AutoDrive-P3 use GRPO, reward-driven optimization, or related alignment strategies to improve reasoning and planning performance~\citep{li2025driver1,zheng2025driveagentr1,fu2025minddrive,zhang2025omnidriver1,ye2026autodrivep3}. Such methods would be difficult or impossible to implement with closed APIs because the optimization requires direct access to model parameters or at least to trainable policy components. In this sense, the move toward open backbones is not only an engineering shift; it is a methodological precondition for reasoning alignment.

A more recent development is the emergence of world-model and VLA-style backbones. Instead of treating language models as the central reasoning engine, some methods use video diffusion models, latent world models, or vision-language-action models to encode physical dynamics more directly~\citep{nvidia2025alpamayor1,bytedance2025uniugp,zheng2025world4drive,luo2026lastvla}. This suggests a deeper infrastructure transition: the backbone of driving CoT is moving from language-centric intelligence toward physically grounded, action-conditioned foundation models. The likely future is not a pure LLM planner wrapped around perception modules, but an integrated stack in which language, vision, dynamics, and action are jointly represented.

The infrastructure shift therefore has a clear implication. Early driving CoT asked whether a general LLM could reason about driving when given a textual description of the scene. Current work increasingly asks how to train, align, and deploy specialized open models whose internal representations are grounded in perception, dynamics, and action. This shift explains why the field has rapidly moved from prompting to fine-tuning, from text-only models to multimodal models, and from language reasoning to action-grounded reasoning.

\subsection{The Cognitive Shift: What Do Models Actually Reason About?}
\label{app:shift-cognitive}

\begin{figure*}[t]
  \centering
  \includegraphics[width=\textwidth]{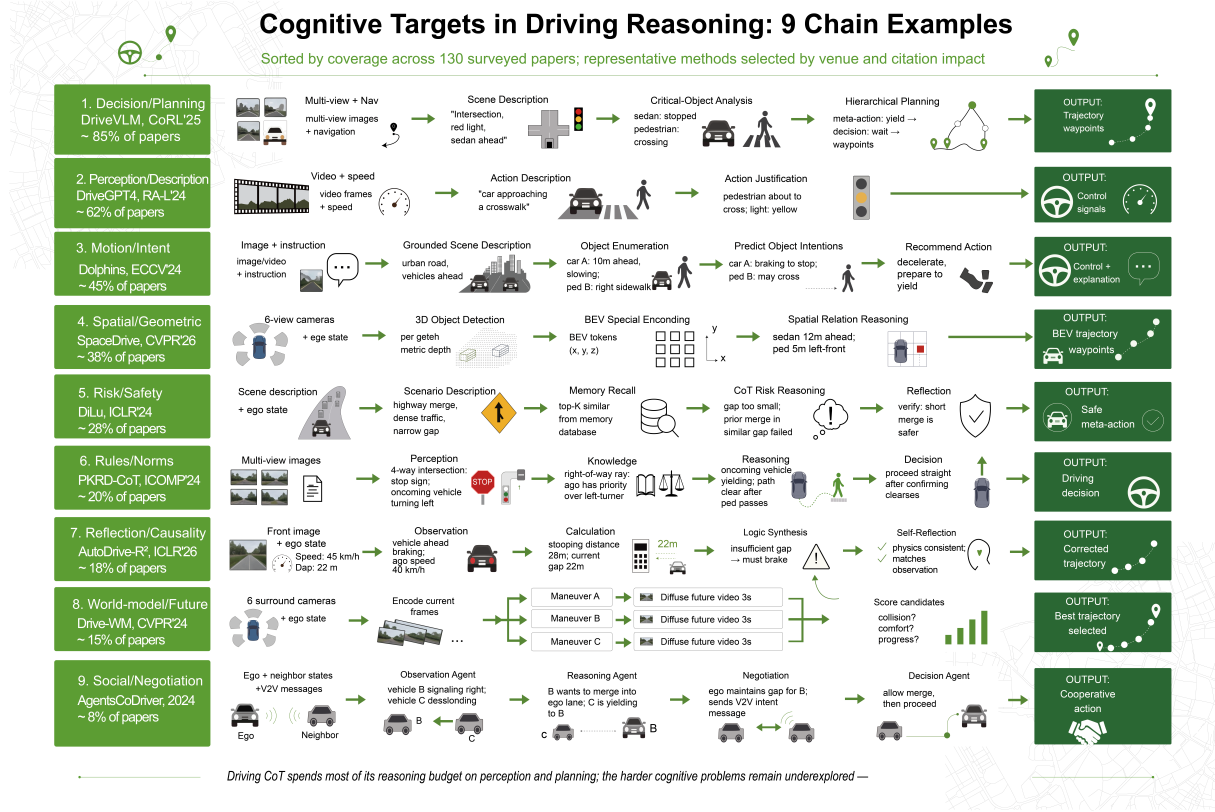}
  \caption{Nine cognitive targets of intermediate reasoning in driving CoT, ordered by frequency. Each strip shows a representative reasoning chain from one method. Strip height is proportional to the number of methods addressing each target, illustrating the concentration on decision/planning and perception tasks and the scarcity of rule reasoning and social negotiation.}
  \label{fig:shift-cognitive}
\end{figure*}

Although many papers claim to perform reasoning, the content of that reasoning varies widely. Some models reason about what is visible in the scene; some reason about future motion; some reason about traffic rules; some reason about risk, uncertainty, or social interaction. To understand this variation, we classify the intermediate reasoning content in the surveyed methods into several cognitive targets, including perception and description, spatial and geometric understanding, motion and intent prediction, decision and planning, risk and safety, traffic rules, reflection and causality, world-model prediction, and social negotiation.

The most common target is decision and planning. Many methods organize their reasoning chain around the sequence from perception to prediction to planning and finally to behavior selection~\citep{wang2023drivemlm,tian2024drivevlm,wang2024drivecot,luo2024pkrdcot,liao2025cotdrive}. This is understandable because autonomous driving ultimately requires an action. A reasoning trace that does not affect maneuver choice, trajectory generation, or control remains incomplete from a driving perspective. Procedural methods such as DriveVLM, DriveCoT, PKRD-CoT, and CoT4AD explicitly structure reasoning around driving-relevant stages rather than free-form explanation~\citep{tian2024drivevlm,wang2024drivecot,luo2024pkrdcot,wang2025cot4ad}.

The second major target is perception and scene description. Methods such as DriveGPT4, Dolphins, EMMA, OpenEMMA, and ReasonDrive use language to describe objects, road context, ego behavior, and scene-level risk factors~\citep{xu2023drivegpt4,ma2023dolphins,hwang2024emma,xing2024openemma,chahe2025reasondrive}. This form of reasoning is useful for interpretability and human interaction, but it also exposes a limitation: describing a scene is not the same as reasoning about how to act. A model may correctly mention a pedestrian, vehicle, or lane boundary while still failing to produce a safe trajectory. Therefore, perception-description reasoning is necessary but not sufficient for action-grounded CoT.

Spatial and geometric reasoning forms another important target. Driving decisions depend on distances, lanes, drivable areas, occlusions, right-of-way geometry, and relative motion. Language-only reasoning is weak at preserving such metric structure, which motivates methods that incorporate BEV features, object-level crops, occupancy maps, or spatially grounded visual prompts~\citep{yang2024driveoccworld,corbiere2025retrievalbased,li2025spacedrive,chen2025solve,min2026visionlanguageaction}. Benchmarks such as NuScenes-SpatialQA also show that spatial reasoning remains a major weakness of current VLMs in driving scenarios~\citep{tian2025nuscenesspatialqa}. This suggests that spatial reasoning should not be treated as a minor subtask of language explanation; it is one of the central bottlenecks for reliable driving CoT.

A smaller but increasingly important group of methods focuses on future-oriented reasoning. World-model-based approaches generate future frames, occupancy states, or latent rollouts to support planning under uncertainty~\citep{wang2023drivewm,gao2024vista,zeng2025futuresightdrive,zheng2025world4drive,xiong2026unidrivewm}. These methods shift reasoning from ``what is happening now'' to ``what may happen next.'' This is a crucial cognitive transition because many driving decisions cannot be made from the current frame alone. Merging, yielding, unprotected turns, occluded pedestrians, and aggressive neighboring vehicles all require forecasting. However, future prediction also introduces the risk of compounding error: a visually plausible future may not be physically or behaviorally useful for planning.

Reflection and causality represent another emerging target. Instead of producing a single reasoning chain, reflective methods revise, critique, or counterfactually evaluate possible decisions~\citep{yuan2025autodriver,li2025reflectdrive,zhang2025openread,peng2025counterfactual,luo2026unleashing}. This changes CoT from a static explanation into an error-correction process. In safety-critical driving, this is especially valuable because the first plan may be unsafe, incomplete, or inconsistent with traffic context. The limitation is cost: reflection usually requires extra inference steps, additional verification, or multiple candidate plans. As a result, reflective reasoning is more suitable for difficult or uncertain scenarios than for every frame.

The least developed cognitive targets are rules, norms, risk, and social negotiation. Yet these are precisely the areas where foundation models could provide the most value beyond traditional perception and planning pipelines. Rule-grounded methods such as PKRD-CoT, DiLu, KoMA, and related agent-based approaches attempt to inject traffic laws, right-of-way reasoning, human preference, or multi-agent interaction into the decision process~\citep{luo2024pkrdcot,wen2024dilu,jiang2024koma,zheng2024planagent,mao2024agent}. Cooperative methods further extend reasoning across vehicles or infrastructure through message exchange~\citep{hu2024agentscodriver,chiu2025v2vllm,liu2025colmdriver,gao2025langcoop,hou2025driveagent}. However, these topics remain underrepresented compared with perception and planning. This imbalance suggests that current driving CoT still spends much of its reasoning budget on tasks that overlap with conventional modules, while the harder cognitive problems of norm interpretation, negotiation, and risk-aware decision-making remain relatively open.

The cognitive shift therefore reveals a deeper issue. The field is moving from descriptive reasoning toward action-oriented reasoning, but not all cognitively important targets have advanced equally. Future driving CoT should place more emphasis on the problems that classical pipelines handle poorly: ambiguous right-of-way, social interaction, rare traffic rules, causal failure analysis, and uncertainty-aware risk reasoning.

\subsection{The Action Shift: From Talking to Doing}
\label{app:shift-action}

\begin{figure*}[t]
  \centering
  \includegraphics[width=\textwidth]{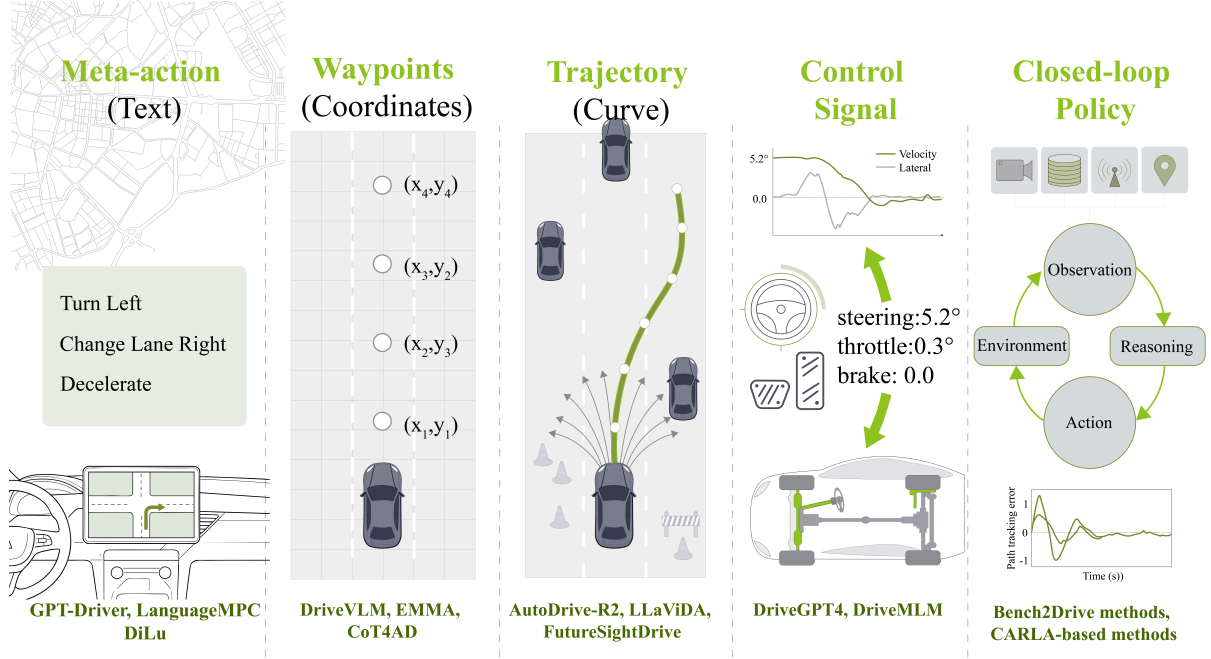}
  \caption{Five forms of action output in driving CoT methods. From left to right: meta-action as high-level text commands, discrete waypoint coordinates in bird's-eye view, continuous trajectory curves, low-level control signals, and closed-loop policies with observation-action feedback. Representative methods are listed below each type.}
  \label{fig:shift-action}
\end{figure*}

The most visible evolution in driving CoT is the shift from textual explanation to trajectory-oriented output. Early methods often produced scene descriptions, driving advice, control commands, or high-level maneuvers~\citep{xu2023drivegpt4,sha2023languagempc,ma2023dolphins}. These outputs were useful for demonstrating that language models could reason about driving scenes, but they did not always close the loop to executable behavior. More recent methods increasingly output waypoints, trajectories, or planner-compatible action representations~\citep{tian2024drivevlm,wang2025cot4ad,zhou2025autovla,zhang2025reasoningvla,ye2026autodrivep3}. This marks an important change: CoT is no longer treated merely as a text-generation mechanism for explanation, but as an intermediate process that should support driving behavior.

However, outputting a trajectory does not automatically mean that the reasoning trace faithfully caused the action. The corpus reveals several forms of weak coupling between reasoning and behavior. The first is parallel or decoupled generation. In some architectures, the model produces a reasoning text and a trajectory from shared visual features, but the trajectory does not explicitly depend on the generated reasoning. In such cases, the text may explain the action after the fact rather than determine it. This creates the risk of post-hoc rationalization, where the reasoning appears coherent but is not causally linked to the final driving decision.

The second weak-coupling pattern is training-only reasoning. Methods such as VERDI, DSDrive, and other distillation-based approaches use explicit textual reasoning during training but compress or remove it at inference time for efficiency~\citep{feng2025verdi,liu2025dsdrive,qiao2025lightemma}. This design is practically motivated: verbose language reasoning is too slow for real-time driving. Yet it raises a critical question. If reasoning is internalized into hidden states, how can we verify that the deployed model preserves the causal and semantic benefits of the original reasoning traces? A faster model may inherit the performance gains of CoT supervision without preserving the faithfulness of the reasoning process.

The third pattern is QA-only reasoning. Driving QA and explanation benchmarks are valuable because they test whether models can answer questions, describe scenes, and explain decisions~\citep{sima2023drivelm,wang2024omnidrive,wei2025driveqa,zhang2025bench2advlm,zeng2025dvbench}. However, QA performance does not guarantee planning performance. A model can correctly answer a question about a scene and still fail to generate a safe trajectory. Conversely, a planner may produce a safe trajectory while giving an incomplete or misleading explanation. This creates a reasoning-planning disconnect that has been observed in recent analysis work~\citep{drivemind2025}.

The fourth pattern is reward-driven decoupling. With the rise of reinforcement fine-tuning (RFT), models can learn to exploit the reward function rather than achieving the true objective — a well-known reward-hacking failure mode amplified by the strong optimization capability of VLMs and the misalignment between the true goal and the reward proxy. In driving, this typically manifests as a decoupling between the CoT trace and the predicted vehicle behavior. Because directly mapping visual inputs to trajectories is more straightforward than predicting a CoT and then conditioning the trajectory on it, the model tends to take the shortcut and ignore its own reasoning when only the trajectory is rewarded. Symmetrically, when the textual CoT is rewarded independently, the model can satisfy the linguistic reward through format manipulation or hallucinated content. Without a consistency penalty linking the two heads, the model's articulated intent, claiming to yield to a vulnerable road user for instance, can directly contradict the trajectory it actually executes.
  
This action shift therefore exposes the central challenge of driving CoT: faithfulness. A useful reasoning representation should not merely accompany an action; it should constrain, guide, or verify the action. Future evaluation should include causal intervention tests. For example, if a critical object is removed from the reasoning trace, the trajectory should change in a predictable way. If irrelevant text is perturbed, the action should remain stable. If a retrieved rule contradicts the scene, the system should either reject it or explain why it is inapplicable. Such tests are necessary because conventional open-loop metrics cannot tell whether the intermediate reasoning actually matters.

The field has made real progress from talking about driving to producing driving actions. But the harder transition is from action output to action-faithful reasoning. A method that outputs waypoints on nuScenes or NAVSIM is still not fully action-grounded if the reasoning trace can be removed, corrupted, or replaced without changing the trajectory. The next stage of driving CoT should therefore move beyond trajectory generation alone and evaluate whether reasoning is causally coupled to closed-loop behavior.

\subsection{The Deployment Shift: The Cost of Thinking}
\label{app:shift-deployment}

\begin{figure*}[t]
  \centering
  \includegraphics[width=\textwidth]{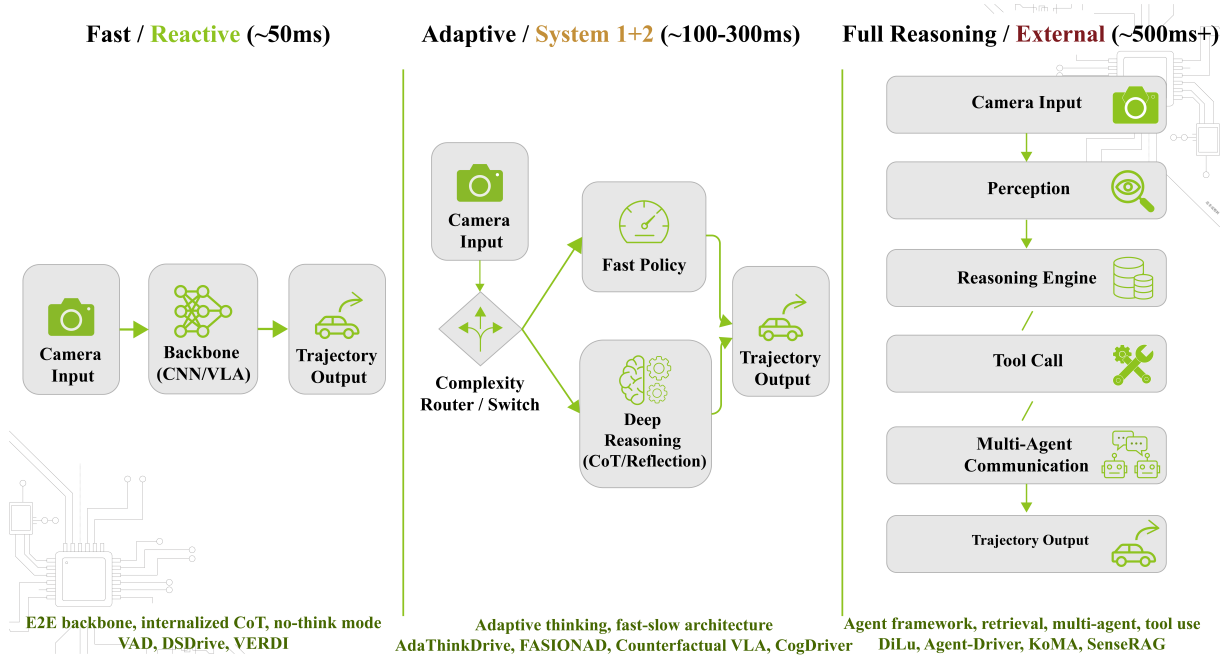}
  \caption{Three deployment strategies for driving CoT, illustrating the trade-off between reasoning depth and inference latency. Left: fast reactive policies that internalize reasoning into a single forward pass (${\sim}$50\,ms). Middle: adaptive systems that route simple scenes to a fast path and complex scenes to a deliberative reasoning path (${\sim}$100--300\,ms). Right: full external reasoning pipelines involving memory, tool calls, and multi-agent communication (${\sim}$500\,ms+).}
  \label{fig:shift-deployment}
\end{figure*}

Reasoning is useful only if it can be deployed under realistic constraints. Autonomous driving requires low latency, temporal consistency, robustness to distribution shift, and safe fallback behavior. These requirements change how CoT should be designed. In general language tasks, longer reasoning may improve accuracy. In driving, however, longer reasoning can be dangerous if it delays action, accumulates errors, or produces unstable decisions across time.

Different reasoning representations face different deployment bottlenecks. Linguistic CoT is highly interpretable but slow because autoregressive text generation introduces substantial latency~\citep{tian2024drivevlm,wang2024drivecot,yuan2025autodriver}. It also struggles with fine-grained spatial grounding, since language is a lossy medium for metric geometry and temporal motion~\citep{tian2025nuscenesspatialqa,godbole2025dramax}. Externalized CoT, including retrieval, tool use, and multi-agent communication, can improve long-tail knowledge and modularity, but it introduces additional delays and interface failures~\citep{luo2025senserag,chang2025drivingrag,qian2025agentthink,chiu2025v2vllm}. Retrieved evidence may be irrelevant, a tool may be called in the wrong context, or a communication message may be delayed or inconsistent.

Visual-spatial reasoning has a different bottleneck. Future frames, occupancy maps, masks, crops, and trajectory sketches can ground reasoning in scene evidence, but they may suffer from error accumulation and incomplete evidence selection~\citep{wang2023drivewm,yang2024driveoccworld,zeng2025futuresightdrive,corbiere2025retrievalbased,li2025spacedrive}. A generated future may look realistic while failing to preserve the interaction variables needed for safe planning. A selected crop may focus on a visible vehicle while missing an occluded pedestrian. Thus, visual reasoning improves grounding but does not automatically guarantee completeness or safety.

Latent-dynamic reasoning is often more compatible with planning and control, but its deployment challenge is verification. Latent rollouts, dynamics tokens, and optimization states are difficult to inspect directly~\citep{zheng2025world4drive,shang2026dynvla,yang2025worldrft,gao2025diffvla}. This creates a blind spot: latent reasoning may be action-effective but hard to certify. If a latent rollout leads to an unsafe decision, it is difficult to determine whether the failure came from perception, dynamics modeling, reward design, or trajectory optimization. For production systems, this lack of interpretability cannot be ignored simply because the representation is internal.

The rise of reinforcement learning introduces another deployment challenge: reward design. Recent methods use GRPO, DPO, or reward-driven optimization to align reasoning and planning~\citep{li2025driver1,zheng2025driveagentr1,zhang2025omnidriver1,fu2025minddrive,ye2026autodrivep3}. Trajectory-level objectives can reward collision avoidance, route progress, comfort, and displacement error. But reasoning quality is harder to reward. A reasoning trace should be grounded, concise, faithful, risk-aware, and causally connected to action. No widely accepted reward captures all of these properties. As a result, RL may improve planning metrics while leaving the reasoning process underconstrained.

These bottlenecks motivate adaptive reasoning. Instead of running full CoT at every frame, several recent methods activate deeper reasoning only when the scene is difficult, uncertain, or risky~\citep{luo2025adathinkdrive,qian2024fasionad,peng2025counterfactual,xie2025deliberation}. Fast-slow architectures use lightweight reactive policies for routine driving and slower deliberative modules for complex scenarios. Counterfactual or reflective systems trigger additional reasoning when the model is uncertain or when candidate plans conflict. Hierarchical systems use high-level language reasoning at low frequency and low-level control at high frequency~\citep{zhang2025lead}. This design better matches the structure of driving: most frames do not require deep reasoning, but rare frames may require substantial deliberation.

The deployment shift therefore changes the central design question. The goal is not to build a vehicle that always reasons more. The goal is to build a vehicle that reasons at the right time, with the right representation, at the right computational cost. In routine scenarios, compact latent or reactive policies may be sufficient. In ambiguous intersections, occluded hazards, rule conflicts, or multi-agent negotiation, deeper language, retrieval, simulation, or tool-mediated reasoning may be necessary. Future driving CoT should therefore be evaluated not only by reasoning accuracy, but also by reasoning efficiency, trigger reliability, fallback safety, and closed-loop stability.

\paragraph{Summary.}
The four shifts described above show that driving CoT is undergoing a transition from language-prompted explanation to action-grounded, trainable, and deployment-aware reasoning. The infrastructure shift enables model adaptation; the cognitive shift expands what counts as reasoning; the action shift demands causal coupling between intermediate representations and behavior; and the deployment shift forces reasoning to become adaptive and cost-aware. Together, these trends support the central claim of this survey: the future of CoT in autonomous driving is not longer textual chains, but intermediate representations that can be grounded, executed, verified, and trusted under real driving constraints.

\section{Full Comparison Table}
\label{app:comparison-table}

Table~\ref{tab:full-comparison} lists the papers in our corpus and groups the
130 method papers according to their primary reasoning representation. The
remaining benchmark, dataset, survey, and analysis papers are listed separately
when applicable. Backbone is reported at the model-family level; ``--''
indicates that the specific model was not identifiable from publicly available
sources. Because many systems combine multiple representations, the category
assigned in the table reflects the representation that plays the most central
role in the reasoning-action pipeline. For hybrid systems, the Subtype column
lists the primary label with a secondary tag in parentheses, where (K) denotes
a knowledge-grounded and (R) a retrieval-augmented secondary role.

\begin{figure}[t]
  \centering
  \includegraphics[width=0.85\linewidth]{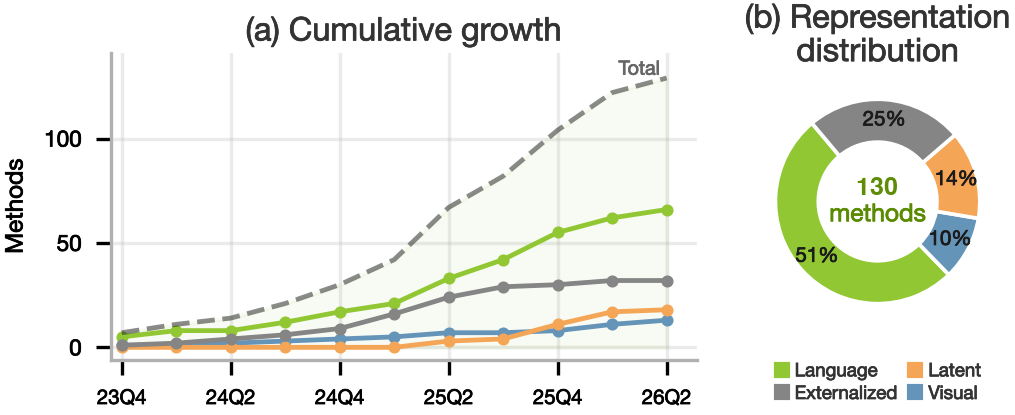}
  \caption{Temporal growth and representation distribution of the 130 method papers. Left: cumulative number of methods by primary reasoning representation from 2023Q4 to 2026Q2. Right: distribution over the four categories.}
  \label{fig:representation-trend}
\end{figure}

\section{Benchmark Landscape}
\label{app:benchmark-details}
This appendix expands the compact discussion in Section~\ref{sec:benchmarks}.
The benchmark literature does not evaluate a single unified notion of
``reasoning.'' Instead, different datasets test different parts of the
reasoning-action pipeline: language explanations, spatial grounding,
robustness, trajectory planning, and closed-loop behavior.

\subsection{Evaluation Focus}

\paragraph{Reasoning and QA benchmarks.}
DriveLM~\citep{sima2023drivelm}, Reason2Drive~\citep{nie2023reason2drive},
DriveLMM-o1~\citep{ishaq2025drivelmmo1},
AD\textsuperscript{2}-Bench~\citep{wei2025adbench},
DriveQA~\citep{wei2025driveqa}, WOMD-Reasoning~\citep{li2025womdreasoning},
DriveCombo~\citep{ma2026drivecombo}, and AgentDrive~\citep{ferrag2026agentdrive}
evaluate whether models can describe driving scenes, answer structured
questions, follow traffic rules, or produce step-wise explanations. These
benchmarks are useful for language-based representations, but most remain
open-loop and do not directly test whether a reasoning trace changes the final
trajectory.

\paragraph{Spatial grounding and robustness benchmarks.}
NuScenes-SpatialQA~\citep{tian2025nuscenesspatialqa},
VLADBench~\citep{li2025finegrained}, DVBench~\citep{zeng2025dvbench},
DRAMA-X~\citep{godbole2025dramax}, and
RoboDriveVLM~\citep{liao2025robodrivevlm} focus on object grounding, spatial
relations, intent prediction, safety-critical perception, and robustness under
corruption. They are especially important for visual-spatial reasoning because
they test whether intermediate representations are grounded in the correct
objects, locations, and relations before action is produced.

\paragraph{Planning and action benchmarks.}
nuScenes~\citep{nuscenes}, nuPlan~\citep{nuplan}, NAVSIM~\citep{navsim},
CoVLA~\citep{sasaki2024covla}, DriveAction~\citep{hao2025driveaction},
doScenes~\citep{martinezsanchez2026doscenes}, and
AutoDriDM~\citep{lian2026autodridm} evaluate trajectory prediction,
planning quality, action prediction, or instruction-conditioned driving
outputs. These benchmarks connect reasoning representations to motion outputs,
but they often report final planning metrics without evaluating whether the
intermediate reasoning is faithful to the action.

\paragraph{Closed-loop benchmarks.}
Bench2Drive~\citep{jia2024bench2drive},
Bench2ADVLM~\citep{zhang2025bench2advlm}, and
Bench2Drive-VL~\citep{jia2026bench2drivevl} place driving models in
interactive simulation or real/sim closed-loop settings. They measure route
completion, driving score, infractions, and safety outcomes, making them the
closest benchmarks to deployment-oriented evaluation. However, even these
benchmarks usually evaluate behavior rather than the correctness, stability, or
faithfulness of the intermediate reasoning representation.

\paragraph{Coverage gaps.}
The current landscape is fragmented. QA benchmarks evaluate reasoning quality
without action consequences; planning benchmarks evaluate actions without
inspecting intermediate reasoning; grounding benchmarks expose perceptual
errors without measuring downstream control impact; and closed-loop benchmarks
rarely measure latency, retrieval reliability, tool failure, or multi-agent
message consistency.  The taxonomy in Section~\ref{sec:taxonomy} defines four reasoning \emph{artifacts}: language, visual-spatial, latent-dynamic, and externalized. While the evaluation literature defines several orthogonal \emph{dimensions} along which any artifact can be tested: (i) grounding correctness, whether the artifact encodes the right scene content; (ii) action coupling, whether the artifact actually drives the produced action; (iii) reasoning cost, whether the artifact is affordable in real-time deployment; (iv) closed-loop reliability, whether the artifact remains stable when actions affect future observations; and (v) faithfulness under intervention, whether perturbing the artifact predictably perturbs the action. A useful way to read the current literature is as a matrix of \emph{artifact $\times$ dimension} cells. Existing benchmarks cluster in a narrow band of this matrix: language artifacts probed for grounding correctness, or arbitrary artifacts scored by open-loop trajectory error. Latent-dynamic representations are barely tested for grounding or faithfulness, because their internal states are not directly readable; externalized representations are rarely tested for source reliability, delay, or conflict handling. Most cells in the matrix are empty, and this emptiness, rather than the absence of any single benchmark, is what makes cross-representation comparison difficult.

\paragraph{Implications.}
This fragmentation suggests that future benchmarks should move from
\emph{task-level scoring} to \emph{representation-conditional evaluation}.
A driving model should be judged not only by whether it answers correctly,
plans safely, or completes a route, but also by whether its intermediate
reasoning artifact is scenario-appropriate and causally useful for the final
action. For language artifacts, this means evaluating object- and rule-level
faithfulness; for visual-spatial artifacts, whether the selected region is
causally relevant to the maneuver; for latent-dynamic artifacts, whether
internal rollouts or state transitions can be probed or perturbed; and for
externalized artifacts, whether source reliability, delay, and conflict
handling are explicitly measured.

\paragraph{From coverage to causality.}
In our opinion, filling the matrix above requires two evaluation primitives that current protocols lack. The first is \emph{intervention-based faithfulness}: perturbing or ablating a part of the reasoning artifact and checking whether the action changes in a scene-consistent way. The probe is naturally representation-specific (swap the object for language, replace the case for retrieval, intervene on the rolled-out state for latent rollouts, inject a known-wrong response for tools), and without it post-hoc rationalization is statistically indistinguishable from causal reasoning. The second is \emph{stratified evaluation by scenario difficulty}: aggregate L2 or collision rates dilute the value of CoT, because reasoning's marginal benefit lies in long-tail cases, not in routine lane following. Reporting reasoning-conditional metrics by difficulty stratum, paired with cost measurements (latency, retrieval calls, tool invocations, communication rounds), turns the matrix from a coverage chart into a causal one and points toward \emph{representation-conditional protocols}, rather than yet another end-to-end driving suite.
Two safeguards apply to such interventions: the perturbation must preserve artifact validity rather than create an implausible or out-of-distribution state, and the expected behavioral change is assessed at the artifact's abstraction level rather than by exact waypoint correspondence. These representation-specific probes are proposed operational diagnostics for evaluating the surveyed systems, not results claimed or run by this survey.

\subsection{Metric Glossary}
\label{app:metric-glossary}

Table~\ref{tab:metric-glossary} summarizes the most common metric families in the evaluation landscape and their main limitations.

\begin{table}[h]
\centering
\small
\setlength{\tabcolsep}{4pt}
\begin{tabular}{@{}p{0.30\linewidth} p{0.30\linewidth} p{0.31\linewidth}@{}}
\toprule
Metric & What it measures & Main limitation \\
\midrule
Exact-match / multiple-choice accuracy & Agreement with a predefined answer & Does not establish grounding or planner use \\
L2 / ADE / FDE & Distance from the logged trajectory at selected or averaged future steps & Penalizes alternative safe futures; definitions and horizons vary \\
Open-loop collision rate & Predicted geometric overlap with obstacles & Protocol-dependent and does not capture interactive reactions \\
Closed-loop or composite score, including PDMS/EPDMS & Progress, compliance, comfort, and infractions under a benchmark protocol & Simulator- and version-specific; aggregate values may hide component failures \\
\bottomrule
\end{tabular}
\caption{Metric glossary for the evaluation landscape.}
\label{tab:metric-glossary}
\end{table}

\subsection{Benchmark Inventory}

Table~\ref{tab:benchmark-details} lists the benchmark, dataset, and analysis
papers used in this survey. Compared with Table~\ref{tab:benchmark-landscape},
which groups benchmarks by evaluation focus, this inventory serves as a
fine-grained appendix-level reference. It records the individual benchmark
purpose, task setting, main output format, and reported metrics whenever they
are identifiable from the paper. Since benchmarks with similar high-level goals
may still evaluate different parts of the reasoning-action pipeline, the
inventory helps readers locate whether each benchmark supports language
rationales, spatial grounding, trajectory quality, robustness, or closed-loop
safety.

\onecolumn

\begin{figure*}[p]
  \centering
  \resizebox{\textwidth}{!}{%
    \input{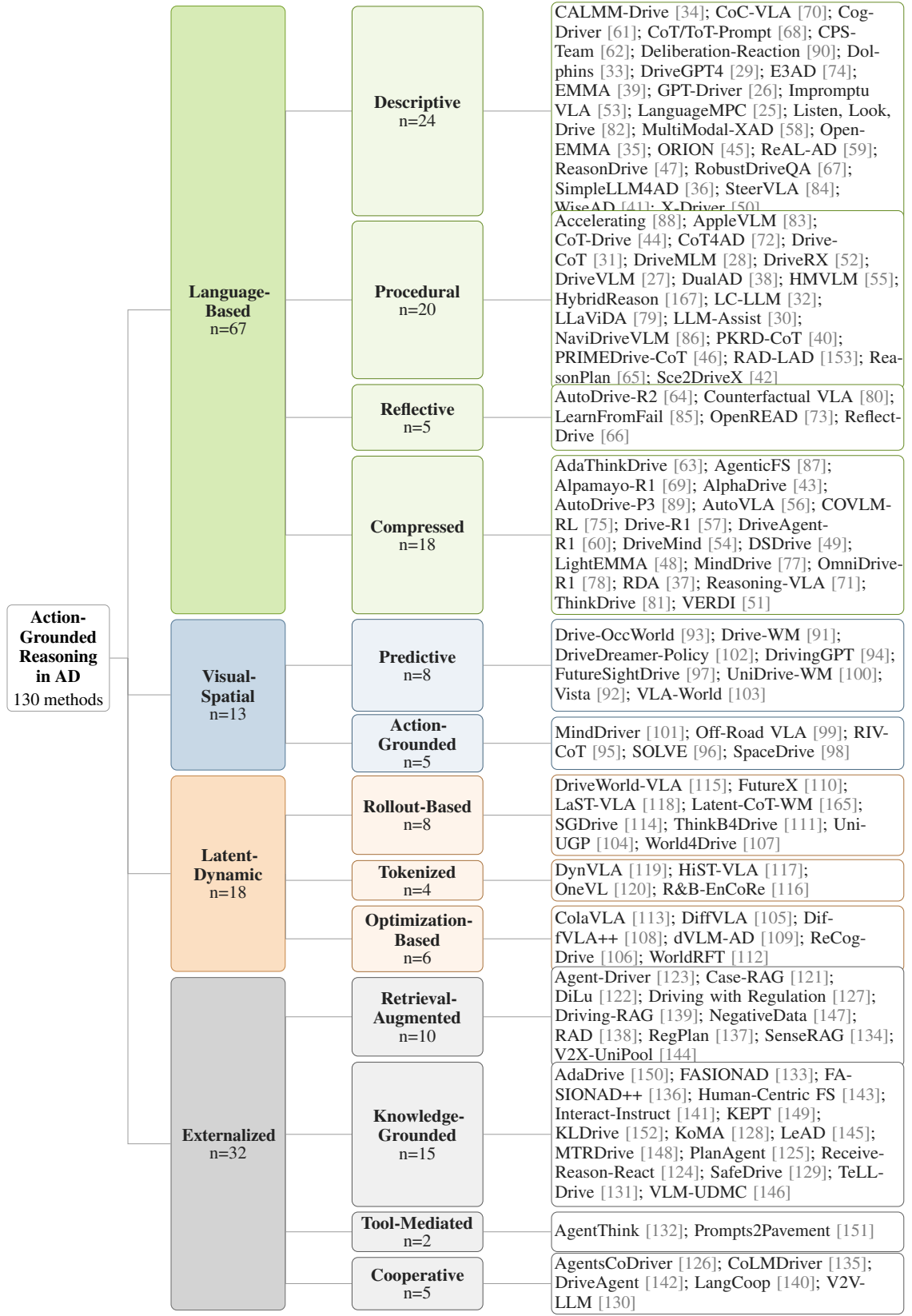}%
  }
  \caption{Full taxonomy of 130 method papers organized by primary intermediate reasoning representation. Numbers in brackets are reference indices.}
  \label{fig:full-taxonomy}
\end{figure*}

\small
\setlength{\tabcolsep}{2pt}
\renewcommand{\arraystretch}{1.02}

\begin{longtable}{>{\raggedright\arraybackslash}p{0.045\textwidth} >{\raggedright\arraybackslash}p{0.31\textwidth} >{\raggedright\arraybackslash}p{0.13\textwidth} >{\raggedright\arraybackslash}p{0.115\textwidth} >{\raggedright\arraybackslash}p{0.095\textwidth} >{\raggedright\arraybackslash}p{0.16\textwidth}}
\caption{Full comparison of method papers, grouped by primary reasoning representation.}
\label{tab:full-comparison} \\
\toprule
Year & Paper & Subtype & Backbone & Training & Evaluation \\
\midrule
\endhead
\midrule \multicolumn{6}{r}{Continued on next page} \\
\endfoot
\bottomrule
\endlastfoot

\midrule
\multicolumn{6}{l}{\textbf{I. Language-based CoT}} \\
\midrule
2024 & Making Large Language Models Better Planners with Reasoning-Decision Alignment \citep{huang2024making} & Compressed & LLaMA & SFT & nuScenes, DriveLM \\
2025 & AlphaDrive \citep{jiang2025alphadrive} & Compressed & Qwen & RL+SFT & MetaAD  \\
2025 & LightEMMA \citep{qiao2025lightemma} & Compressed & Qwen+LLaMA & Prompt & nuScenes \\
2025 & DSDrive \citep{liu2025dsdrive} & Compressed & Qwen+LLaMA & Distill & CARLA \\
2025 & VERDI \citep{feng2025verdi} & Compressed & Qwen & Distill & nuScenes,HugSim \\
2025 & DriveMind \citep{wasif2025drivemind} & Compressed & CLIP-based & RL+Distill & CARLA \\
2025 & AutoVLA \citep{zhou2025autovla} & Compressed & Qwen & RL+SFT & nuScenes, NAVSIM, Waymo E2E, Bench2Drive \\
2025 & Drive-R1 \citep{li2025driver1} & Compressed & InternVL & RL+SFT & nuScenes, DriveLM \\
2025 & DriveAgent-R1 \citep{zheng2025driveagentr1} & Compressed & Qwen & RL+SFT & nuScenes \\
2025 & AdaThinkDrive \citep{luo2025adathinkdrive} & Compressed & InternVL & RL+SFT & NAVSIM \\
2025 & Alpamayo-R1 (AR1) \citep{nvidia2025alpamayor1} & Compressed & Cosmos-Reason & RL+SFT & open-loop + closed-loop \\
2025 & Reasoning-VLA \citep{zhang2025reasoningvla} & Compressed & Qwen & RL+SFT & nuScenes, NAVSIM \\
2025 & COVLM-RL \citep{li2025covlmrl} & Compressed & InternVL & RL+Prompt & CARLA \\
2025 & MindDrive \citep{fu2025minddrive} & Compressed & Qwen & RL+SFT & Bench2Drive \\
2025 & OmniDrive-R1 \citep{zhang2025omnidriver1} & Compressed & Qwen & RL & DriveLMM-o1 \\
2026 & ThinkDrive \citep{zhao2026thinkdrive} & Compressed & Qwen& RL+SFT & DrivingVQA \\
2026 & Bridging Large-Model Reasoning \citep{chen2026bridging} & Compressed & InternVL+DeepSeek & SFT+Prompt & CARLA \\
2026 & AutoDrive-P3 \citep{ye2026autodrivep3} & Compressed & Qwen & RL+SFT & nuScenes, NAVSIM \\
2023 & GPT-Driver \citep{mao2023gptdriver} & Descriptive & GPT & SFT & nuScenes \\
2023 & DriveGPT4 \citep{xu2023drivegpt4} & Descriptive & LLaMA & SFT & BDD-X \\
2023 & LanguageMPC \citep{sha2023languagempc} & Descriptive & GPT & Prompt & IDSim \\
2023 & Dolphins \citep{ma2023dolphins} & Descriptive & OpenFlamingo & instruction tuning & BDD-X \\
2024 & SimpleLLM4AD \citep{zheng2024simplellm4ad} & Descriptive & InternVL & SFT & DriveLM-nuScenes \\
2024 & EMMA \citep{hwang2024emma} & Descriptive & Gemini & SFT & nuScenes, Waymo \\
2024 & CALMM-Drive \citep{yao2024calmmdrive} & Descriptive & GPT & Prompt & nuPlan \\
2024 & WiseAD \citep{zhang2024wisead} & Descriptive & LLaMA & SFT & CARLA \\
2024 & OpenEMMA \citep{xing2024openemma} & Descriptive & Qwen+LLaMA & Prompt & nuScenes \\
2025 & ORION \citep{fu2025orion} & Descriptive & Vicuna & SFT & Bench2Drive \\
2025 & ReasonDrive \citep{chahe2025reasondrive} & Descriptive & Qwen+LLaMA & SFT+Distill & DriveLM \\
2025 & X-Driver \citep{liu2025xdriver} & Descriptive & LLaVA & SFT & CARLA,Bench2Drive \\
2025 & Impromptu VLA \citep{chi2025impromptu} & Descriptive & Qwen & SFT+Distill & nuScenes,NeuroNCAP\\
2025 & Multimodal Framework for Explainable Autonomous Driving \citep{zarghani2025multimodal} & Descriptive & VideoMAE  & SFT & nuScenes, BDD-X \\
2025 & ReAL-AD \citep{lu2025realad} & Descriptive & LLaMA+Qwen & SFT & nuScenes, Bench2Drive \\
2025 & OmniReason \citep{liu2025omnireason} & Descriptive & LLaVA & SFT+Distill & nuScenes,Bench2Drive \\
2025 & The System Description of CPS  \citep{peng2025system} & Descriptive & LLaVA & SFT & DriveLM-nuScenes \\
2025 & Robust Driving QA through Meta \citep{yu2025robust} & Descriptive & Qwen & Prompt & driving QA benchmark \\
2025 & Enhancing Vision-Language Mode \citep{wu2025enhancing} & Descriptive & Qwen & Prompt & RoboSense Challenge at IROS 2025 \\
2025 & CoC-VLA \citep{zhang2025cocvla} & Descriptive & LLaVA & SFT & nuScenes + CARLA \\
2025 & E3AD \citep{liao2025e3ad} & Descriptive & Qwen & SFT & Talk2Car, DrivePilot, MoCAD,Talk2Car-Trajectory \\
2026 & Listen, Look, Drive \citep{guo2026listen} & Descriptive & Qwen & SFT & nuScenes \\
2026 & SteerVLA \citep{gao2026steervlab} & Descriptive & InternVL & SFT & Bench2Drive \\
2026 & Deliberation Meets Reaction \citep{xie2025deliberation} & Descriptive & InternVL & SFT & Bench2Drive \\
2023 & DriveMLM \citep{wang2023drivemlm} & Procedural & LLaMA & SFT & CARLA \\
2024 & LLM-Assist \citep{sharan2024llmassist} & Procedural & LLaMA+GPT & Prompt & nuPlan \\
2024 & DriveVLM \citep{tian2024drivevlm} & Procedural & Qwen & SFT & nuScenes \\
2024 & DriveCoT \citep{wang2024drivecot} & Procedural & VideoSwin-Transformer & SFT + Distill & CARLA \\
2024 & LC-LLM \citep{peng2024lcllm} & Procedural & LLaMA & SFT & highD\\
2024 & DualAD \citep{wang2024dualad} & Procedural & GLM+GPT & Prompt & nuPlan \\
2024 & PKRD-CoT \citep{luo2024pkrdcot} & Procedural & Qwen+GPT & Prompt & nuScenes \\
2025 & Sce2DriveX \citep{zhao2025sce2drivex} & Procedural & Vicuna & SFT & nuScenes, Bench2Drive \\
2025 & CoT-Drive \citep{liao2025cotdrive} & Procedural & GPT & Distill & NGSIM,
HighD, MoCAD, ApolloScape,nuScenes \\
2025 & PRIMEDrive-CoT \citep{mandalika2025primedrivecot} & Procedural & Bayesian Graph Neural Network & SFT & DriveCoT \\
2025 & DriveRX \citep{diao2025driverx} & Procedural & DeepSeek & RL & DriveBench, DriveLM-Hard \\
2025 & HMVLM \citep{wang2025hmvlm} & Procedural & Qwen-VL & SFT & Waymo \\
2025 & ReasonPlan \citep{liu2025reasonplan} & Procedural & Qwen & SFT+Self-sup & Bench2Drive \\
2025 & CoT4AD \citep{wang2025cot4ad} & Procedural & LLaMA & SFT & nuScenes, Bench2Drive \\
2025 & LLaViDA \citep{liu2025llavida} & Procedural & LLaVA & SFT & nuScenes \\
2026 & AppleVLM \citep{han2026applevlm} & Procedural & Janus Pro & SFT & CARLA \\
2026 & NaviDriveVLM \citep{tao2026navidrivevlm} & Procedural & Qwen-VL & SFT & nuScenes \\
2026 & RAD-LAD \citep{ghosh2026radlad} & Procedural & Qwen & 3-stage curriculum& nuPlan \\
2026 & Accelerating Structured Chain-of-Thought in Autonomous Vehicles \citep{gu2026accelerating} & Procedural & Qwen & SFT & Internal \\
2024 & Hybrid Reasoning Based on Large Language Models for Autonomous Car Driving \citep{azarafza2024hybrid} & Procedural (K) & GPT & Prompt & CARLA \\
2025 & AutoDrive-R\textsuperscript{2} \citep{yuan2025autodriver} & Reflective & Qwen & RL+SFT & nuScenes, NAVSIM, Waymo \\
2025 & Discrete Diffusion for Reflective Vision-Language-Action Models in Autonomous Driving \citep{li2025reflectdrive} & Reflective & LLaDA-V & SFT & NAVSIM \\
2025 & OpenREAD \citep{zhang2025openread} & Reflective & Qwen & RL+SFT & nuScenes \\
2025 & Counterfactual VLA \citep{peng2025counterfactual} & Reflective & Qwen & SFT & Internal \\
2026 & Unleashing VLA Potentials in Autonomous Driving via Explicit Learning from Failures \citep{luo2026unleashing} & Reflective & Qwen+InternVL & RL+SFT & NAVSIM \\
\midrule
\multicolumn{6}{l}{\textbf{II. Visual CoT}} \\
\midrule
2025 & Retrieval-Based Interleaved Visual Chain-of-Thought in Real-World Driving Scenarios \citep{corbiere2025retrievalbased} & Action-grounded & Qwen+LLaVA & SFT & DrivingVQA \\
2025 & SOLVE \citep{chen2025solve} & Action-grounded & LLaMA& SFT & nuScenes \\
2025 & SpaceDrive \citep{li2025spacedrive} & Action-grounded & Qwen & SFT & nuScenes, Bench2Drive \\
2026 & A Vision-Language-Action Model with Visual Prompt for OFF-Road Autonomous Driving \citep{min2026visionlanguageaction} & Action-grounded & Qwen & Prompt & RELLIS-3D (off-road). \\
2026 & MindDriver \citep{zhang2026minddriver} & Action-grounded & Qwen & RL+SFT & nuScenes, CARLA, Bench2Drive \\
2023 & Driving into the Future \citep{wang2023drivewm} & Predictive & World/Diff & Diffusion & nuScenes \\
2024 & Vista \citep{gao2024vista} & Predictive & World/Diff & Diffusion & nuScenes, OpenDV \\
2024 & Driving in the Occupancy World \citep{yang2024driveoccworld} & Predictive & BEVFormer-style & SFT & nuScenes \\
2024 & DrivingGPT \citep{chen2024drivinggpt} & Predictive & LLaMA & SFT & NAVSIM, nuPlan \\
2025 & FutureSightDrive \citep{zeng2025futuresightdrive} & Predictive & Qwen & unified pretraining  & nuScenes, NAVSIM, DriveLM \\
2026 & UniDrive-WM \citep{xiong2026unidrivewm} & Predictive & Vicuna & SFT & nuScenes, Bench2Drive \\
2026 & DriveDreamer-Policy \citep{zhou2026drivedreamerpolicy} & Predictive & Qwen & SFT & NAVSIM \\
2026 & Learning Vision-Language-Action World Models for Autonomous Driving \citep{wang2026vlaworld} & Predictive & Qwen & RL+SFT & nuScenes \\
\midrule
\multicolumn{6}{l}{\textbf{III. Latent / Simulative CoT}} \\
\midrule
2025 & DiffVLA \citep{jiang2025diffvla} & Optimization-based & LLaMA & SFT & NAVSIM \\
2025 & ReCogDrive \citep{li2025recogdrive} & Optimization-based & Qwen + InternVL & GRPO + SFT & NAVSIM, CARLA, Bench2Drive \\
2025 & DiffVLA++ \citep{gao2025diffvla} & Optimization-based & LLaMA & SFT & NAVSIM \\
2025 & dVLM-AD \citep{ma2025dvlmad} & Optimization-based & SigLIP-based & SFT & nuScenes, Waymo \\
2025 & WorldRFT \citep{yang2025worldrft} & Optimization-based & E2E CNN/Trans & RL+SFT & nuScenes, NAVSIM \\
2025 & ColaVLA \citep{peng2025colavla} & Optimization-based & LLaMA& SFT & nuScenes \\
2025 & UniUGP \citep{bytedance2025uniugp} & Rollout-based & Qwen & SFT  & nuScenes, Waymo, DriveLM \\
2025 & World4Drive \citep{zheng2025world4drive} & Rollout-based & E2E CNN/Trans & SFT & nuScenes, NAVSIM \\
2025 & FutureX \citep{lin2025futurex} & Rollout-based & model-agnostic& SFT & NAVSIM, CARLA \\
2025 & Latent Chain-of-Thought World Modeling for End-to-End Driving \citep{latentcotwm2025} & Rollout-based & Qwen & RL+SFT & PhysicalAI-AV\\
2025 & Think Before You Drive \citep{liao2025think} & Rollout-based & BERT & SFT & Talk2Car, MoCAD, DrivePilot, RefCOCO/+/g \\
2026 & SGDrive \citep{li2026sgdrive} & Rollout-based & InternVL & RL+SFT & NAVSIM \\
2026 & DriveWorld-VLA \citep{jia2026driveworldvla} & Rollout-based & InternVL & SFT & nuScenes, NAVSIM \\
2026 & LaST-VLA \citep{luo2026lastvla} & Rollout-based & InternVL & RL+SFT & NAVSIM \\
2026 & R\&B-EnCoRe \citep{ganai2026rbencore} & Tokenized & Qwen+LLaMA & 2-stage variational  & LIBERO-90, nuScenes \\
2026 & HiST-VLA \citep{wang2026histvla} & Tokenized & LLaMA & SFT & NAVSIM \\
2026 & DynVLA \citep{shang2026dynvla} & Tokenized & EMU3 & RL+SFT & NAVSIM, CARLA, Bench2Drive \\
2026 & OneVL \citep{team2026onevl} & Tokenized & Qwen & SFT & NAVSIM \\
\midrule
\multicolumn{6}{l}{\textbf{IV. Externalized / Distributed CoT}} \\
\midrule
2024 & AgentsCoDriver \citep{hu2024agentscodriver} & Cooperative & GPT & Prompt & highway-env multi-vehicle \\
2025 & V2V-LLM \citep{chiu2025v2vllm} & Cooperative & LLaVA & SFT & V2V-QA \\
2025 & CoLMDriver \citep{liu2025colmdriver} & Cooperative & InternVL & SFT+Prompt & CARLA \\
2025 & LangCoop \citep{gao2025langcoop} & Cooperative & Qwen+LLaMA & Prompt & CARLA \\
2025 & DriveAgent \citep{hou2025driveagent} & Cooperative & LLaMA & SFT & Internal \\
2023 & Receive, Reason, and React \citep{cui2023receive} & Knowledge-grounded & GPT & Prompt & HighwayEnv \\
2024 & PlanAgent \citep{zheng2024planagent} & Knowledge-grounded & GPT & Prompt & nuPlan \\
2024 & KoMA \citep{jiang2024koma} & Knowledge-grounded & GPT & Prompt & HighwayEnv \\
2024 & FASIONAD \citep{qian2024fasionad} & Knowledge-grounded & Qwen & SFT & nuScenes, CARLA \\
2024 & SafeDrive \citep{zhou2024safedrive} & Knowledge-grounded & GPT & Prompt & HighD, InD \\
2025 & TeLL-Drive \citep{xu2025telldrive} & Knowledge-grounded & GPT & RL+SFT & HighwayEnv \\
2025 & FASIONAD++ \citep{qian2025fasionad} & Knowledge-grounded & Qwen & SFT & nuScenes, CARLA, Bench2Drive \\
2025 & Interact, Instruct to Improve \citep{fang2025interact} & Knowledge-grounded & LLaMA & Prompt & simulated heterogeneous HVs + real world \\
2025 & Towards Human-Centric Autonomous Driving \citep{xu2025humancentric} & Knowledge-grounded & GPT & RL+Prompt & HighwayEnv \\
2025 & LeAD \citep{zhang2025lead} & Knowledge-grounded & Qwen & Prompt & CARLA \\
2025 & VLM-UDMC \citep{liu2025vlmudmc} & Knowledge-grounded & GPT & SFT+Prompt & simulation + real-world \\
2025 & MTRDrive \citep{luo2025mtrdrive} & Knowledge-grounded & Qwen & RL+SFT & NAVSIM \\
2025 & KEPT \citep{wang2025kept} & Knowledge-grounded & Qwen & SFT & nuScenes \\
2025 & AdaDrive \citep{zhang2025adadrive} & Knowledge-grounded & LLaMA & SFT & LangAuto \\
2026 & KLDrive \citep{tian2026kldrive} & Knowledge-grounded & Qwen & Prompt & nuScenes \\
2024 & Driving with Regulation \citep{cai2024regulation} & Retrieval-augmented & GPT & Prompt & DriveReg \\
2024 & DiLu \citep{wen2024dilu} & Retrieval-augmented (K) & GPT & Prompt & HighwayEnv \\
2024 & A Language Agent for Autonomous Driving \citep{mao2024agent} & Retrieval-augmented (K) & GPT & SFT & nuScenes \\
2025 & SenseRAG \citep{luo2025senserag} & Retrieval-augmented & LLaMA & Prompt & real-world V2X \\
2025 & Traffic Regulation-aware Path  \citep{han2025traffic} & Retrieval-augmented & LLaVA & Prompt & simulation + real-world \\
2025 & RAD \citep{wang2025rad} & Retrieval-augmented & Qwen & SFT & nuScenes \\
2025 & Driving-RAG \citep{chang2025drivingrag} & Retrieval-augmented & GPT & Prompt & CitySim \\
2025 & V2X-UniPool \citep{luo2025v2xunipool} & Retrieval-augmented & GPT & Prompt & DAIR-V2X (real-world V2X) \\
2025 & Case-based Reasoning Augmented \citep{gan2025casebased} & Retrieval-augmented & LLaMA & Prompt & Internal \\
2025 & The Case for Negative Data \citep{patrikar2025case} & Retrieval-augmented & GPT+Mistral & Prompt & nuScenes \\
2025 & AgentThink \citep{qian2025agentthink} & Tool-mediated & Qwen & RL+SFT & DriveLMM-o1 \\
2026 & From Prompts to Pavement \citep{goba2026prompts} & Tool-mediated & GPT & Prompt & CARLA \\

\end{longtable}
\normalsize

\begin{figure}[t]
\centering
\includegraphics[width=\textwidth]{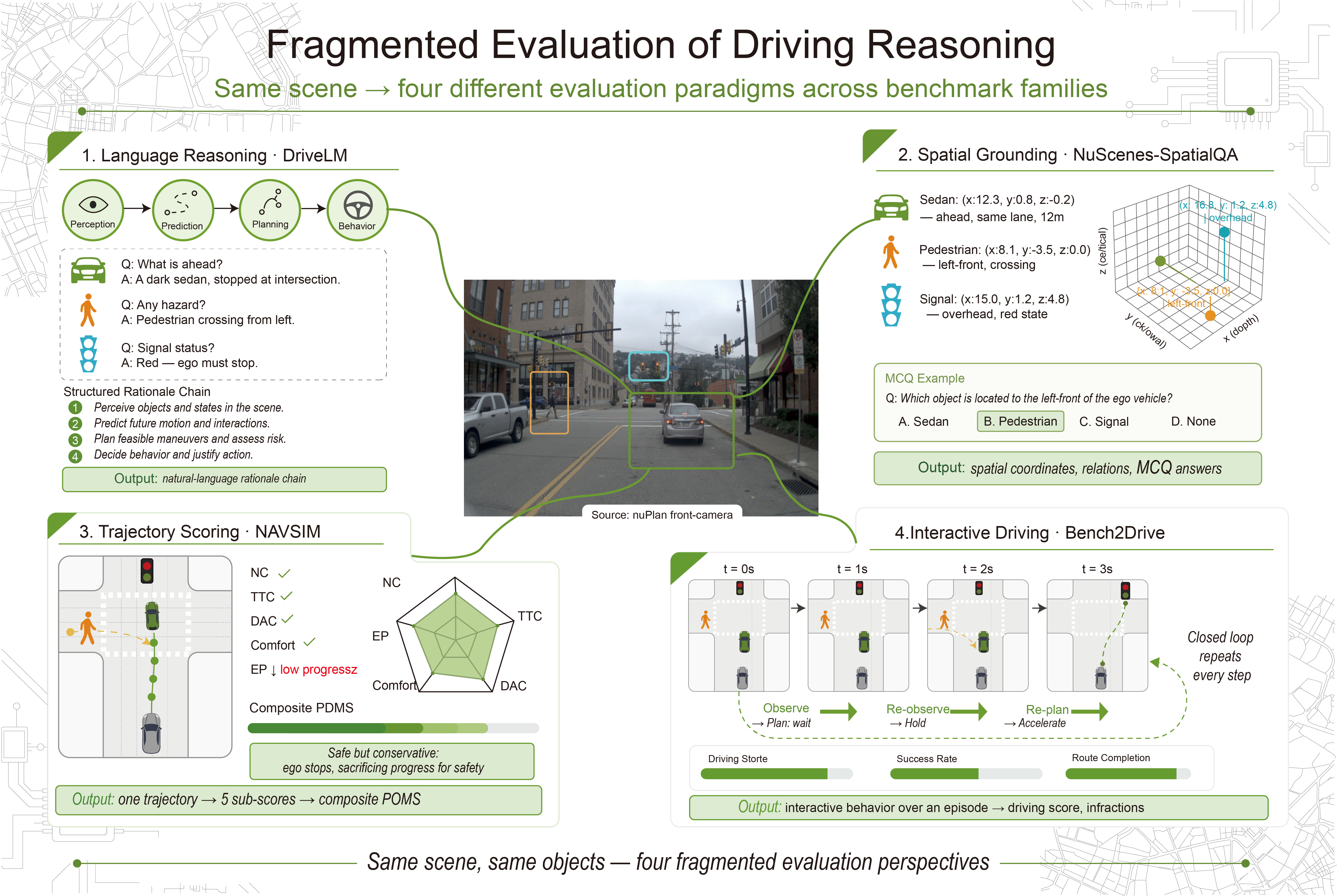}
\caption{Fragmented evaluation of driving reasoning across benchmark families. A single nuPlan front-camera scene with three annotated objects (sedan, pedestrian, traffic light) is processed into four fundamentally different evaluation targets. \textbf{Top-left}: DriveLM converts objects into a Graph-VQA rationale chain across perception, prediction, planning, and behavior stages. \textbf{Top-right}: NuScenes-SpatialQA converts the same objects into 3D coordinates and pairwise spatial relation queries. \textbf{Bottom-left}: NAVSIM projects objects as geometric constraints in a BEV trajectory scoring framework with PDMS sub-score decomposition. \textbf{Bottom-right}: Bench2Drive instantiates objects as interactive agents in CARLA closed-loop simulation. Colored lines trace each object from the center image to its representation in each panel, illustrating how no single benchmark jointly evaluates reasoning quality, spatial grounding, trajectory faithfulness, and closed-loop safety.}

\label{fig:benchmark-landscape}
\end{figure}

\small
\setlength{\tabcolsep}{2.5pt}
\renewcommand{\arraystretch}{1.05}

\begin{longtable}{p{0.14\textwidth} p{0.23\textwidth} p{0.14\textwidth} p{0.18\textwidth} p{0.16\textwidth} p{0.13\textwidth}}
\caption{Benchmark landscape for action-grounded reasoning in autonomous driving.}
\label{tab:benchmark-landscape} \\
\toprule
Focus & Representative benchmarks & Setting & Main output & Primary metrics & Best used for \\
\midrule
\endfirsthead
\toprule
Focus & Representative benchmarks & Setting & Main output & Primary metrics & Best used for \\
\midrule
\endhead
\bottomrule
\endlastfoot

Reasoning / QA &
DriveLM~\citep{sima2023drivelm}; Reason2Drive~\citep{nie2023reason2drive}; DriveLMM-o1~\citep{ishaq2025drivelmmo1}; AD\textsuperscript{2}-Bench~\citep{wei2025adbench}; DriveQA~\citep{wei2025driveqa}; DriveCombo~\citep{ma2026drivecombo}; AgentDrive~\citep{ferrag2026agentdrive} &
Mostly open-loop QA &
Answers, rationales, rule decisions, staged reasoning &
Accuracy, reasoning score, per-step correctness, language metrics &
Language-based reasoning and rule reasoning \\

Spatial grounding &
NuScenes-SpatialQA~\citep{tian2025nuscenesspatialqa}; VLADBench~\citep{li2025finegrained}; DVBench~\citep{zeng2025dvbench}; DRAMA-X~\citep{godbole2025dramax} &
Open-loop perception / QA &
Object grounding, spatial relations, intent, risk labels &
Spatial accuracy, detection, intent accuracy, MCQ accuracy &
Visual-spatial grounding and perception-faithfulness checks \\

Robustness &
RoboDriveVLM~\citep{liao2025robodrivevlm}; DVBench~\citep{zeng2025dvbench}; AD\textsuperscript{2}-Bench~\citep{wei2025adbench} &
Open-loop perturbation tests &
Predictions or answers under corruption, weather, or safety-critical cases &
Trajectory error, collision, robustness drop, adverse-condition accuracy &
Failure modes under distribution shift \\

Planning / trajectory &
nuScenes~\citep{nuscenes}; nuPlan~\citep{nuplan}; NAVSIM~\citep{navsim}; CoVLA~\citep{sasaki2024covla}; DriveAction~\citep{hao2025driveaction}; doScenes~\citep{martinezsanchez2026doscenes} &
Open-loop planning and action prediction &
Waypoints, trajectories, actions, instruction-conditioned plans &
L2/ADE, collision, PDMS/EPDMS, action accuracy &
Action coupling and trajectory quality \\

Closed-loop driving &
Bench2Drive~\citep{jia2024bench2drive}; Bench2ADVLM~\citep{zhang2025bench2advlm}; Bench2Drive-VL~\citep{jia2026bench2drivevl}; nuPlan~\citep{nuplan} &
Closed-loop simulation or real/sim evaluation &
Interactive driving behavior over routes or scenarios &
Driving score, success rate, route completion, infractions &
Closed-loop reliability and safety consequences \\

\end{longtable}
\normalsize

\small
\setlength{\tabcolsep}{2.5pt}
\renewcommand{\arraystretch}{1.05}

\begin{longtable}{p{0.19\textwidth} p{0.055\textwidth} p{0.14\textwidth} p{0.25\textwidth} p{0.16\textwidth} p{0.16\textwidth}}
\caption{Individual benchmark and dataset inventory.}
\label{tab:benchmark-details} \\
\toprule
Benchmark & Year & Focus & What it evaluates & Setting & Main metrics \\
\midrule
\endfirsthead
\toprule
Benchmark & Year & Focus & What it evaluates & Setting & Main metrics \\
\midrule
\endhead
\bottomrule
\endlastfoot

Reason2Drive~\citep{nie2023reason2drive} & 2023 & Reasoning / QA & Chain-based driving reasoning across perception, prediction, and planning questions & Open-loop QA & Aggregated reasoning score; answer accuracy \\
Evaluation of Large Language Models~\citep{inoue2023evaluation} & 2023 & Rule reasoning & Spatially aware decision-making and traffic-rule compliance with text-prompted LLMs & Open-loop decision tasks & Collision avoidance; rule-compliance accuracy \\
DriveLM~\citep{sima2023drivelm} & 2023 & Reasoning / QA & Graph-structured visual QA over perception, prediction, planning, and behavior stages & Open-loop Graph-VQA & QA accuracy; BLEU/ROUGE/CIDEr; planning-oriented QA \\
OmniDrive~\citep{wang2024omnidrive} & 2024 & Multimodal QA / planning & 3D-aware vision-language alignment with DriveLM-style QA and planning outputs & Open-loop QA + nuScenes planning & QA accuracy; L2; collision \\
Bench2Drive~\citep{jia2024bench2drive} & 2024 & Closed-loop driving & Interactive closed-loop driving across disentangled CARLA scenarios & Closed-loop simulation & Driving score; route completion; infractions; success rate \\
CoVLA~\citep{sasaki2024covla} & 2024 & VLA dataset & Paired vision-language-action data for trajectory and language-conditioned driving & Open-loop trajectory + language & Trajectory error; caption / language quality \\
DriveLMM-o1~\citep{ishaq2025drivelmmo1} & 2025 & Step-wise reasoning & Step-wise visual reasoning faithfulness across perception, prediction, and planning & Open-loop VQA & Final-answer accuracy; reasoning-quality score \\
VLADBench~\citep{li2025finegrained} & 2025 & Fine-grained QA & Closed-form QA from traffic knowledge to scene and decision reasoning & Open-loop QA & Per-domain and per-aspect accuracy \\
NuScenes-SpatialQA~\citep{tian2025nuscenesspatialqa} & 2025 & Spatial grounding & Qualitative and quantitative spatial reasoning in driving scenes & Open-loop QA & Spatial accuracy; distance / relation correctness \\
DVBench~\citep{zeng2025dvbench} & 2025 & Safety reasoning & VLLM perception and reasoning over safety-critical driving situations & Open-loop MCQ & Accuracy over hierarchical ability categories \\
WOMD-Reasoning~\citep{li2025womdreasoning} & 2025 & Interaction reasoning & Interaction, traffic-rule, and trajectory-intent reasoning over driving scenes & Open-loop QA + prediction & QA quality; interaction-prediction accuracy \\
DriveAction~\citep{hao2025driveaction} & 2025 & Action coupling & Action-rooted vision-language-action coupling and modality necessity & Open-loop action prediction & Action accuracy; modality ablations \\
AD\textsuperscript{2}-Bench~\citep{wei2025adbench} & 2025 & CoT evaluation & CoT reasoning quality of MLLMs under adverse weather and complex scenes & Open-loop CoT evaluation & Per-step CoT accuracy; end-to-end CoT correctness \\
Bench2ADVLM~\citep{zhang2025bench2advlm} & 2025 & Closed-loop ADVLM & Closed-loop behavior of autonomous-driving VLM agents in simulation and physical tests & Closed-loop sim + real & Driving score; safety score; stage-wise closed-loop metrics \\
DRAMA-X~\citep{godbole2025dramax} & 2025 & Grounding / intent & Object detection and directional intent prediction for risk reasoning & Open-loop multi-task evaluation & Detection mAP; intent accuracy; risk reasoning score \\
DriveQA~\citep{wei2025driveqa} & 2025 & Rule QA & Rule, sign, and right-of-way knowledge for LLMs and MLLMs & Open-loop QA & Per-category accuracy \\
Reasoning-Planning Disconnect~\citep{drivemind2025} & 2025 & Faithfulness analysis & Whether language reasoning causally improves downstream planning quality & Open-loop with nuPlan metrics & Planning scores under CoT removal / intervention \\
RoboDriveVLM~\citep{liao2025robodrivevlm} & 2025 & Robustness & Robustness of VLM-based trajectory prediction under sensor and visual corruptions & Open-loop corruption tests & Trajectory error; collision under corruption \\
From Segments to Scenes~\citep{alvar2025segments} & 2025 & Scenario dataset & Natural-language scene construction from driving segments & Dataset / benchmark & Scene-level coverage; segment-to-scene quality \\
AutoDriDM~\citep{lian2026autodridm} & 2026 & Decision-centric QA & Perception-to-decision ability boundary across object, scene, and decision levels & Open-loop QA & Per-dimension accuracy; perception-decision consistency \\
AgentDrive~\citep{ferrag2026agentdrive} & 2026 & Agentic reasoning & Agentic LLM reasoning across physics, policy, hybrid, and safety dimensions & Open-loop MCQ + simulation & Per-dimension MCQ accuracy; simulation safety \\
doScenes~\citep{martinezsanchez2026doscenes} & 2026 & Instruction planning & Natural-language instruction-conditioned trajectory planning quality & Open-loop trajectory prediction & ADE; instruction-ablation metrics \\
DriveCombo~\citep{ma2026drivecombo} & 2026 & Compositional rules & Compositional multi-rule traffic reasoning under conflicting rules & Open-loop QA & Per-level accuracy; rule-conflict resolution \\
Bench2Drive-VL~\citep{jia2026bench2drivevl} & 2026 & Closed-loop VLM4AD & Closed-loop VLM4AD behavior with auto-generated behavior questions & Closed-loop driving + QA & Driving score; planning / QA metrics \\

\end{longtable}
\normalsize
\newpage
\section{Reported Results by Benchmark}
\label{app:reported-results}

The following tables aggregate reported scores from the surveyed method papers. Metrics are grouped by benchmark family and metric type. Values within a metric group may be sorted by their reported direction, but different metric groups should not be compared as a single leaderboard.

\subsection*{nuScenes}

\scriptsize
\setlength{\tabcolsep}{1pt}
\renewcommand{\arraystretch}{1.08}

\begin{longtable}{>{\raggedright\arraybackslash}p{0.09\textwidth} >{\raggedright\arraybackslash}p{0.16\textwidth} >{\raggedright\arraybackslash}p{0.10\textwidth} >{\raggedright\arraybackslash}p{0.22\textwidth} >{\raggedright\arraybackslash}p{0.055\textwidth} >{\raggedright\arraybackslash}p{0.29\textwidth}}
\caption{Reported results for nuScenes grouped by metric type.}
\label{tab:reported-nuscenes} \\
\toprule
Metric group & Method & Representation & Metric & Value & Protocol / Notes \\
\midrule
\endfirsthead
\toprule
Metric group & Method & Representation & Metric & Value & Protocol / Notes \\
\midrule
\endhead
\midrule \multicolumn{6}{r}{Continued on next page} \\
\endfoot
\bottomrule
\endlastfoot

\midrule
\multicolumn{6}{l}{\textbf{Planning error}} \\
\midrule
Planning error & DriveAgent-R1 \citep{zheng2025driveagentr1} & Compressed & ADE Avg & 0.28 & nuScenes; open-loop; validation\\
Planning error & LightEMMA-QWen \citep{qiao2025lightemma} & Compressed & ADE Avg & 1.45 & nuScenes; open-loop; test \\
Planning error & LightEMMA-LLaMa \citep{qiao2025lightemma} & Compressed & ADE Avg & 1.53 & nuScenes; open-loop; test \\
Planning error & LightEMMA-QWen \citep{qiao2025lightemma} & Compressed & FDE & 2.90 & nuScenes; open-loop; test; 3s \\
Planning error & LightEMMA-LLaMa \citep{qiao2025lightemma} & Compressed & FDE & 3.03 & nuScenes; open-loop; test; 3s \\
Planning error & AutoDrive-R\textsuperscript{2}-7B \citep{yuan2025autodriver} & Reflective & L2 Avg & 0.19 & nuScenes; open-loop; validation \\
Planning error & Reasoning-VLA-7B+ \citep{zhang2025reasoningvla} & Compressed & L2 Avg & 0.22 & nuScenes; open-loop; validation \\
Planning error & Reasoning-VLA-7B \citep{zhang2025reasoningvla} & Compressed & L2 Avg & 0.23 & nuScenes; open-loop; validation \\
Planning error & VLA-World* \citep{wang2026vlaworld} & Predictive & L2 Avg & 0.26 & nuScenes; open-loop; ST-P3 metrics; validation \\
Planning error & SOLVE-VLM \citep{chen2025solve} & Action-grounded & L2 Avg & 0.28 & nuScenes; open-loop; validation \\
Planning error & FSDrive-Qwen-2B* \citep{zeng2025futuresightdrive} & Predictive & L2 Avg & 0.28 & nuScenes; ST-P3 metrics; validation \\
Planning error & ColaVLA \citep{peng2025colavla} & Optimization-based & L2 Avg & 0.30 & nuScenes; open-loop; validation \\
Planning error & VLA-World \citep{wang2026vlaworld} & Predictive & L2 Avg & 0.30 & nuScenes; open-loop; ST-P3 metrics; validation \\
Planning error & Reasoning-VLA-3B \citep{zhang2025reasoningvla} & Compressed & L2 Avg & 0.30 & nuScenes; open-loop; validation \\
Planning error & LLaViDA \citep{liu2025llavida} & Procedural & L2 Avg & 0.31 & nuScenes; open-loop; ST-P3 metrics; validation \\
Planning error & Drive-R1 \citep{li2025driver1} & Compressed & L2 Avg & 0.31 & nuScenes; open-loop; validation \\
Planning error & DriveVLM-Dual \citep{tian2024drivevlm} & Procedural & L2 Avg & 0.31 & nuScenes; validation \\
Planning error & FSDrive-LLaVA-7B* \citep{zeng2025futuresightdrive} & Predictive & L2 Avg & 0.31 & nuScenes; ST-P3 metrics; validation \\
Planning error & SpaceDrive+ \citep{li2025spacedrive} & Action-grounded & L2 Avg & 0.32 & nuScenes; ST-P3 metrics; validation \\
Planning error & AutoDrive-P3-Detailed \citep{ye2026autodrivep3} & Compressed & L2 Avg & 0.33 & nuScenes; open-loop; validation \\
Planning error & MindDriver* \citep{zhang2026minddriver} & Action-grounded & L2 Avg & 0.33 & nuScenes; ST-P3 metrics; validation \\
Planning error & AutoDrive-P3-Fast \citep{ye2026autodrivep3} & Compressed & L2 Avg & 0.34 & nuScenes; open-loop; validation \\
Planning error & DriveVLM \citep{tian2024drivevlm} & Procedural & L2 Avg & 0.40 & nuScenes; validation \\
Planning error & OpenREAD \citep{zhang2025openread} & Reflective & L2 Avg & 0.40 & nuScenes; open-loop; ST-P3 metrics; validation \\
Planning error & VLA-World* \citep{wang2026vlaworld} & Predictive & L2 Avg & 0.42 & nuScenes; open-loop; UniAD metrics; validation \\
Planning error & Drive-OccWorld \citep{yang2024driveoccworld} & Predictive & L2 Avg & 0.47 & nuScenes; TemAvg protocol; validation \\
Planning error & ReAL-AD \citep{lu2025realad} & Descriptive & L2 Avg & 0.48 & nuScenes; ST-P3 metrics; validation \\
Planning error & WorldRFT \citep{yang2025worldrft} & Optimization-based & L2 Avg & 0.48 & nuScenes; open-loop planning; validation \\
Planning error & AutoDrive-R\textsuperscript{2}-3B \citep{yuan2025autodriver} & Reflective & L2 Avg & 0.49 & nuScenes; open-loop; validation \\
Planning error & World4Drive \citep{zheng2025world4drive} & Rollout-based & L2 Avg & 0.50 & nuScenes; open-loop planning; validation \\
Planning error & MindDriver \citep{zhang2026minddriver} & Action-grounded & L2 Avg & 0.53 & nuScenes; ST-P3 metrics; validation \\
Planning error & FSDrive-Qwen-2B \citep{zeng2025futuresightdrive} & Predictive & L2 Avg & 0.53 & nuScenes; ST-P3 metrics; validation \\
Planning error & EchoVLA \citep{guo2026listen} & Descriptive & L2 Avg & 0.58 & nuScenes; validation \\
Planning error & DriveWorld-VLA \citep{jia2026driveworldvla} & Rollout-based & L2 Avg & 0.61 & nuScenes; open-loop planning; validation \\
Planning error & MindDriver* \citep{zhang2026minddriver} & Action-grounded & L2 Avg & 0.65 & nuScenes; open-loop; UniAD metrics; validation \\
Planning error & VERDI \citep{feng2025verdi} & Compressed & L2 Avg & 0.65 & nuScenes; open-loop; validation \\
Planning error & AutoVLA \citep{zhou2025autovla} & Compressed & L2 Avg & 0.70 & nuScenes; UniAD metrics; validation \\
Planning error & Drive-WM \citep{wang2023drivewm} & Predictive & L2 Avg & 0.80 & nuScenes; tree-based planning; validation \\
Planning error & RDA-Driver \citep{huang2024making} & Compressed & L2 Avg & 0.80 & nuScenes; UniAD metrics; open-loop \\
Planning error & VLA-World \citep{wang2026vlaworld} & Predictive & L2 Avg & 0.83 & nuScenes; open-loop; UniAD metrics; validation \\
Planning error & MindDriver \citep{zhang2026minddriver} & Action-grounded & L2 Avg & 0.93 & nuScenes; open-loop; UniAD metrics; validation \\
Planning error & UniUGP \citep{bytedance2025uniugp} & Rollout-based & L2 Avg & 1.23 & nuScenes; planning; front-camera only; validation \\
Planning error & SpaceDrive \citep{li2025spacedrive} & Action-grounded & L2 Avg & 1.8 & nuScenes; open-loop; ST-P3 metrics; validation \\
\midrule
\multicolumn{6}{l}{\textbf{Safety}} \\
\midrule
Safety & WorldRFT \citep{yang2025worldrft} & Optimization-based & Collision Avg & 0.05 & nuScenes; open-loop planning; validation \\
Safety & AutoDrive-P3-Detailed \citep{ye2026autodrivep3} & Compressed & Collision Avg & 0.06 & nuScenes; open-loop; validation \\
Safety & AutoDrive-R\textsuperscript{2}-7B \citep{yuan2025autodriver} & Reflective & Collision Avg & 0.07 & nuScenes; open-loop; validation \\
Safety & Reasoning-VLA-7B+ \citep{zhang2025reasoningvla} & Compressed & Collision Avg & 0.07 & nuScenes; open-loop; validation \\
Safety & AutoDrive-P3-Fast \citep{ye2026autodrivep3} & Compressed & Collision Avg & 0.08 & nuScenes; open-loop; validation \\
Safety & Reasoning-VLA-7B \citep{zhang2025reasoningvla} & Compressed & Collision Avg & 0.08 & nuScenes; open-loop; validation \\
Safety & Agent-Driver \citep{mao2024agent} & Knowledge & Collision Avg & 0.09 & nuScenes; open-loop; ST-P3 metrics; validation \\
Safety & Drive-R1 \citep{li2025driver1} & Compressed & Collision Avg & 0.09 & nuScenes; open-loop; validation \\
Safety & FSDrive-Qwen-2B* \citep{zeng2025futuresightdrive} & Predictive & Collision Avg & 0.10 & nuScenes; ST-P3 metrics; validation \\
Safety & SpaceDrive+ \citep{li2025spacedrive} & Action-grounded & Collision Avg & 0.10 & nuScenes; ST-P3 metrics; validation \\
Safety & VLA-World \citep{wang2026vlaworld} & Predictive & Collision Avg & 0.10 & nuScenes; open-loop; ST-P3 metrics; validation \\
Safety & Drive-OccWorld \citep{yang2024driveoccworld} & Predictive & Collision Avg & 0.11 & nuScenes; TemAvg protocol; validation \\
Safety & OpenREAD \citep{zhang2025openread} & Reflective & Collision Avg & 0.11 & nuScenes; open-loop; ST-P3 metrics; validation \\
Safety & FSDrive-LLaVA-7B* \citep{zeng2025futuresightdrive} & Predictive & Collision Avg & 0.12 & nuScenes; ST-P3 metrics; validation \\
Safety & MindDriver* \citep{zhang2026minddriver} & Action-grounded & Collision Avg & 0.12 & nuScenes; ST-P3 metrics; validation \\
Safety & Reasoning-VLA-3B \citep{zhang2025reasoningvla} & Compressed & Collision Avg & 0.13 & nuScenes; open-loop; validation \\
Safety & DriveAgent-R1 \citep{zheng2025driveagentr1} & Compressed & Collision Avg & 0.14 & nuScenes; open-loop; validation \\
Safety & DriveWorld-VLA \citep{jia2026driveworldvla} & Rollout-based & Collision Avg & 0.16 & nuScenes; open-loop planning; validation \\
Safety & World4Drive \citep{zheng2025world4drive} & Rollout-based & Collision Avg & 0.16 & nuScenes; open-loop planning; validation \\
Safety & FSDrive-Qwen-2B \citep{zeng2025futuresightdrive} & Predictive & Collision Avg & 0.17 & nuScenes; ST-P3 metrics; validation \\
Safety & SOLVE-VLM \citep{chen2025solve} & Action-grounded & Collision Avg & 0.20 & nuScenes; open-loop planning; validation \\
Safety & Agent-Driver \citep{mao2024agent} & Knowledge & Collision Avg & 0.21 & nuScenes; open-loop; UniAD metrics; validation \\
Safety & ColaVLA \citep{peng2025colavla} & Optimization-based & Collision Avg & 0.23 & nuScenes; open-loop planning; validation \\
Safety & Drive-WM \citep{wang2023drivewm} & Predictive & Collision Avg & 0.26 & nuScenes; tree-based planning; validation \\
Safety & SOLVE-E2E \citep{chen2025solve} & Action-grounded & Collision Avg & 0.30 & nuScenes; open-loop planning; validation \\
Safety & AutoVLA \citep{zhou2025autovla} & Compressed & Collision Avg & 0.31 & nuScenes; UniAD metrics; validation \\
Safety & RDA-Driver \citep{huang2024making} & Compressed & Collision Avg & 0.32 & nuScenes; UniAD metrics; open-loop \\
Safety & dVLM-AD \citep{ma2025dvlmad} & Optimization-based & Collision Avg & 0.32 & nuScenes; validation \\
Safety & UniUGP \citep{bytedance2025uniugp} & Rollout-based & Collision Avg & 0.33 & nuScenes; planning; front-camera only; validation \\
Safety & SpaceDrive \citep{li2025spacedrive} & Action-grounded & Collision Avg & 0.76 & nuScenes; open-loop; ST-P3 metrics; validation \\
\midrule
\multicolumn{6}{l}{\textbf{Generation / occupancy}} \\
\midrule
Generation / occupancy & Drive-OccWorld-A \citep{yang2024driveoccworld} & Predictive & mIoU\_\allowbreak{}f & 36.3 & nuScenes; inflated GMO and flow forecasting; validation; 2s \\
Generation / occupancy & Drive-OccWorld-P \citep{yang2024driveoccworld} & Predictive & mIoU\_\allowbreak{}f & 36.3 & nuScenes; inflated GMO and flow forecasting; validation; 2s \\
Generation / occupancy & Drive-OccWorld-P \citep{yang2024driveoccworld} & Predictive & mIoU\_\allowbreak{}f & 21.2 & nuScenes; fine-grained GSO forecasting; validation; 2s \\
Generation / occupancy & Drive-OccWorld-P \citep{yang2024driveoccworld} & Predictive & VPQ\_\allowbreak{}f & 25.1 & nuScenes; inflated GMO and flow forecasting; validation; 2s \\
Generation / occupancy & Drive-OccWorld-A \citep{yang2024driveoccworld} & Predictive & VPQ\_\allowbreak{}f & 23.7 & nuScenes; inflated GMO and flow forecasting; validation; 2s \\
Generation / occupancy & Vista \citep{gao2024vista} & Predictive & FID & 6.9 & nuScenes; prediction fidelity; validation \\
Generation / occupancy & UniUGP \citep{bytedance2025uniugp} & Rollout-based & FID & 7.4 & nuScenes; future frame generation; front-camera only; validation \\
Generation / occupancy & MindDriver \citep{zhang2026minddriver} & Action-grounded & FID & 9.4 & nuScenes; future frame generation; validation \\
Generation / occupancy & VLA-World \citep{wang2026vlaworld} & Predictive & FID & 9.8 & nuScenes; future frame generation; validation \\
Generation / occupancy & FSDrive \citep{zeng2025futuresightdrive} & Predictive & FID & 10.1 & nuScenes; future frame generation; validation \\
Generation / occupancy & Drive-WM \citep{wang2023drivewm} & Predictive & FID & 15.8 & nuScenes; multiview video generation; validation \\
Generation / occupancy & UniUGP \citep{bytedance2025uniugp} & Rollout-based & FVD & 75.9 & nuScenes; future frame generation; front-camera only; validation \\
Generation / occupancy & Vista \citep{gao2024vista} & Predictive & FVD & 89.4 & nuScenes; prediction fidelity; validation \\
Generation / occupancy & Drive-WM \citep{wang2023drivewm} & Predictive & FVD & 122.7 & nuScenes; multiview video generation; validation \\
\midrule
\multicolumn{6}{l}{\textbf{Auxiliary}} \\
\midrule
Auxiliary & VLA-World \citep{wang2026vlaworld} & Predictive & Action F1 (forward) & 95.88 & nuScenes; action prediction; validation \\
Auxiliary & dVLM-AD \citep{ma2025dvlmad} & Optimization-based & Behavior-Trajectory alignment avg & 87.8 & nuScenes; validation \\
Auxiliary & VLA-World \citep{wang2026vlaworld} & Predictive & Action F1 (right) & 75.06 & nuScenes; action prediction; validation \\
Auxiliary & VLA-World \citep{wang2026vlaworld} & Predictive & Action F1 (left) & 74.22 & nuScenes; action prediction; validation \\
Auxiliary & KLDrive \citep{tian2026kldrive} & Knowledge & NuScenes-QA overall accuracy & 65.04 & nuScenes; NuScenes-QA; validation \\
Auxiliary & DriveAgent-R1 \citep{zheng2025driveagentr1} & Compressed & First-frame joint accuracy & 52.96 & nuScenes; high-level planning; test; 8s \\
Auxiliary & DriveAgent-R1 \citep{zheng2025driveagentr1} & Compressed & Sequence average joint accuracy & 47.1 & nuScenes; high-level planning; test; 8s \\
Auxiliary & The Case for Negative Data \citep{patrikar2025case} & Retrieval & VLM-only recall on REASONABLE actions (f\_\allowbreak{}base) & 24 & nuScenes-derived custom benchmark; 1,275 action-scene pairs \\
Auxiliary & VERDI \citep{feng2025verdi} & Compressed & FPS & 4.5 & nuScenes; open-loop; validation \\
Auxiliary & AutoVLA \citep{zhou2025autovla} & Compressed & Runtime (s) & 3.95 & nuScenes; open-loop; test \\
Auxiliary & VERDI \citep{feng2025verdi} & Compressed & Latency (ms) & 359 & HugSim; closed-loop; overall \\
Auxiliary & VERDI \citep{feng2025verdi} & Compressed & COM & 0.954 & HugSim; closed-loop; overall \\
Auxiliary & VERDI \citep{feng2025verdi} & Compressed & DAC & 0.944 & HugSim; closed-loop; overall \\
Auxiliary & VERDI \citep{feng2025verdi} & Compressed & NC & 0.659 & HugSim; closed-loop; overall \\
Auxiliary & VERDI \citep{feng2025verdi} & Compressed & TTC & 0.469 & HugSim; closed-loop; overall \\
Auxiliary & VERDI \citep{feng2025verdi} & Compressed & Rc & 0.354 & HugSim; closed-loop; overall \\
Auxiliary & VERDI \citep{feng2025verdi} & Compressed & HDScore & 0.2 & HugSim; closed-loop; overall 
\end{longtable}
\normalsize

\subsection*{NAVSIM}

\scriptsize
\setlength{\tabcolsep}{1pt}
\renewcommand{\arraystretch}{1.08}

\begin{longtable}{>{\raggedright\arraybackslash}p{0.09\textwidth} >{\raggedright\arraybackslash}p{0.16\textwidth} >{\raggedright\arraybackslash}p{0.10\textwidth} >{\raggedright\arraybackslash}p{0.22\textwidth} >{\raggedright\arraybackslash}p{0.055\textwidth} >{\raggedright\arraybackslash}p{0.29\textwidth}}
\caption{Reported results for NAVSIM grouped by metric type.}
\label{tab:reported-navsim} \\
\toprule
Metric group & Method & Representation & Metric & Value & Protocol / Notes \\
\midrule
\endfirsthead
\toprule
Metric group & Method & Representation & Metric & Value & Protocol / Notes \\
\midrule
\endhead
\midrule \multicolumn{6}{r}{Continued on next page} \\
\endfoot
\bottomrule
\endlastfoot

\midrule
\multicolumn{6}{l}{\textbf{Primary planning scores}} \\
\midrule
Primary planning scores & ReflectDrive† \citep{li2025reflectdrive} & Reflective & PDMS & 94.7 & NAVSIM v1; navtest; oracle reflection upper bound \\
Primary planning scores & AdaThinkDrive-BoN \citep{luo2025adathinkdrive} & Compressed & PDMS & 93.0 & NAVSIM v1; navtest; Best-of-4 oracle selection \\
Primary planning scores & AutoVLA-BoN \citep{zhou2025autovla} & Compressed & PDMS & 92.12 & NAVSIM v1; navtest; Best-of-6 oracle selection; 4s \\
Primary planning scores & DriveWorld-VLA \citep{jia2026driveworldvla} & Rollout-based & PDMS & 91.3 & NAVSIM v1; navtest \\
Primary planning scores & LaST-VLA-8B-RL \citep{luo2026lastvla} & Rollout-based & PDMS & 91.3 & NAVSIM v1; navtest; RL fine-tuned \\
Primary planning scores & ReflectDrive \citep{li2025reflectdrive} & Reflective & PDMS & 91.1 & NAVSIM v1; navtest \\
Primary planning scores & SGDrive-RFT \citep{li2026sgdrive} & Rollout-based & PDMS & 91.1 & NAVSIM v1; navtest; RFT \\
Primary planning scores & ELF-VLA-8B \citep{luo2026unleashing} & Reflective & PDMS & 91.0 & NAVSIM v1; navtest \\
Primary planning scores & ReCogDrive \citep{li2025recogdrive} & Optimization-based & PDMS & 90.8 & NAVSIM v1; navtest \\
Primary planning scores & FutureX-All-TransFuser \citep{lin2025futurex} & Rollout-based & PDMS & 90.6 & NAVSIM v1; navtest; TransFuser backbone (camera+LiDAR) \\
Primary planning scores & AutoDrive-P3-Detailed \citep{ye2026autodrivep3} & Compressed & PDMS & 90.6 & NAVSIM v1; navtest \\
Primary planning scores & AdaThinkDrive \citep{luo2025adathinkdrive} & Compressed & PDMS & 90.3 & NAVSIM v1; navtest; adaptive CoT switching \\
Primary planning scores & AutoDrive-P3-Fast \citep{ye2026autodrivep3} & Compressed & PDMS & 90.2 & NAVSIM v1; navtest \\
Primary planning scores & FutureX-Auto-LTF \citep{lin2025futurex} & Rollout-based & PDMS & 89.2 & NAVSIM v1; navtest; LTF backbone (camera-only) \\
Primary planning scores & DriveDreamer-Policy \citep{zhou2026drivedreamerpolicy} & Predictive & PDMS & 89.2 & NAVSIM v1; navtest \\
Primary planning scores & AutoVLA \citep{zhou2025autovla} & Compressed & PDMS & 89.11 & NAVSIM v1; navtest; SFT+CoT; 4s \\
Primary planning scores & AutoDrive-R\textsuperscript{2}-7B \citep{yuan2025autodriver} & Reflective & PDMS & 90.3 & NAVSIM v1; navtest \\
Primary planning scores & MTRDrive \citep{luo2025mtrdrive} & Knowledge & PDMS & 88.3 & NAVSIM v1; navtest (split not explicitly stated) \\
Primary planning scores & WorldRFT \citep{yang2025worldrft} & Optimization-based & PDMS & 87.8 & NAVSIM v1; navtest; closed-loop planning \\
Primary planning scores & SGDrive-SFT \citep{li2026sgdrive} & Rollout-based & PDMS & 87.4 & NAVSIM v1; navtest; SFT \\
Primary planning scores & World4Drive \citep{zheng2025world4drive} & Rollout-based & PDMS & 85.1 & NAVSIM v1; navtest; closed-loop planning \\
Primary planning scores & AutoDrive-P3-Detailed \citep{ye2026autodrivep3} & Compressed & EPDMS & 89.9 & NAVSIM v2; navtest \\
Primary planning scores & AutoDrive-P3-Fast \citep{ye2026autodrivep3} & Compressed & EPDMS & 88.7 & NAVSIM v2; navtest \\
Primary planning scores & DriveDreamer-Policy \citep{zhou2026drivedreamerpolicy} & Predictive & EPDMS & 88.7 & NAVSIM v2; navtest \\
Primary planning scores & HiST-VLA \citep{wang2026histvla} & Tokenized & EPDMS & 88.6 & NAVSIM v2; navtest \\
Primary planning scores & ELF-VLA-8B \citep{luo2026unleashing} & Reflective & EPDMS & 87.1 & NAVSIM v2; navtest \\
Primary planning scores & LaST-VLA-8B-RL \citep{luo2026lastvla} & Rollout-based & EPDMS & 87.1 & NAVSIM v2; navtest; RL fine-tuned \\
Primary planning scores & DriveWorld-VLA \citep{jia2026driveworldvla} & Rollout-based & EPDMS & 86.8 & NAVSIM v2; navtest \\
Primary planning scores & AutoDrive-P3-Detailed \citep{ye2026autodrivep3} & Compressed & EPDMS & 86.2 & NAVSIM v2; navtest; alt. config \\
Primary planning scores & AutoDrive-P3-Fast \citep{ye2026autodrivep3} & Compressed & EPDMS & 85.2 & NAVSIM v2; navtest; alt. config \\
Primary planning scores & HiST-VLA \citep{wang2026histvla} & Tokenized & EPDMS & 50.9 & NAVSIM v2; Navhard pseudo-closed-loop \\
Primary planning scores & DiffVLA++ \citep{gao2025diffvla} & Optimization-based & EPDMS & 49.12 & NAVSIM v2; ICCV 2025 AGC public leaderboard; combined ensemble \\
Primary planning scores & DiffVLA++-VLA \citep{gao2025diffvla} & Optimization-based & EPDMS & 48.0 & NAVSIM v2; Navhard two-stage test; VLA branch \\
Primary planning scores & DiffVLA \citep{jiang2025diffvla} & Optimization-based & EPDMS & 45.01 & NAVSIM v2; AGC private test; combined \\
Primary planning scores & DiffVLA++-E2E \citep{gao2025diffvla} & Optimization-based & EPDMS & 43.7 & NAVSIM v2; Navhard two-stage test; E2E branch \\
\midrule
\multicolumn{6}{l}{\textbf{Subscores}} \\
\midrule
Subscores & ReflectDrive$^\dagger$ \citep{li2025reflectdrive} & Reflective & NC & 99.7 & NAVSIM v1; navtest \\
Subscores & HiST-VLA \citep{wang2026histvla} & Tokenized & NC & 99.6 & NAVSIM v2; navtest \\
Subscores & AutoDrive-P3-Detailed \citep{ye2026autodrivep3} & Compressed & NC & 99.1 & NAVSIM v1; navtest \\
Subscores & ELF-VLA-8B \citep{luo2026unleashing} & Reflective & NC & 98.9 & NAVSIM v1; navtest \\
Subscores & ELF-VLA-8B \citep{luo2026unleashing} & Reflective & NC & 98.9 & NAVSIM v2; navtest \\
Subscores & SGDrive \citep{li2026sgdrive} & Rollout-based & NC & 98.6 & NAVSIM v1; navtest RFT \\
Subscores & AutoDrive-R\textsuperscript{2}-7B \citep{yuan2025autodriver} & Reflective & NC & 98.5 & NAVSIM v1; navtest \\
Subscores & AdaThinkDrive \citep{luo2025adathinkdrive} & Compressed & NC & 98.4 & NAVSIM v1; navtest \\
Subscores & ReCogDrive \citep{li2025recogdrive} & Optimization-based & NC & 97.9 & NAVSIM v1; navtest \\
Subscores & ReflectDrive \citep{li2025reflectdrive} & Reflective & NC & 97.7 & NAVSIM v1; navtest \\
Subscores & ReflectDrive$^\dagger$ \citep{li2025reflectdrive} & Reflective & DAC & 99.5 & NAVSIM v1; navtest \\
Subscores & ReflectDrive \citep{li2025reflectdrive} & Reflective & DAC & 99.3 & NAVSIM v1; navtest \\
Subscores & HiST-VLA \citep{wang2026histvla} & Tokenized & DAC & 99.1 & NAVSIM v2; navtest \\
Subscores & ELF-VLA-8B \citep{luo2026unleashing} & Reflective & DAC & 98.1 & NAVSIM v1; navtest \\
Subscores & ELF-VLA-8B \citep{luo2026unleashing} & Reflective & DAC & 98.1 & NAVSIM v2; navtest \\
Subscores & AdaThinkDrive \citep{luo2025adathinkdrive} & Compressed & DAC & 97.8 & NAVSIM v1; navtest \\
Subscores & SGDrive \citep{li2026sgdrive} & Rollout-based & DAC & 97.8 & NAVSIM v1; navtest RFT \\
Subscores & AutoDrive-P3-Detailed \citep{ye2026autodrivep3} & Compressed & DAC & 97.4 & NAVSIM v1; navtest \\
Subscores & ReCogDrive \citep{li2025recogdrive} & Optimization-based & DAC & 97.3 & NAVSIM v1; navtest \\
Subscores & AutoVLA-BoN \citep{zhou2025autovla} & Compressed & DAC & 97.08 & NAVSIM v1; open-loop; test; 4s \\
Subscores & AutoDrive-R\textsuperscript{2}-7B \citep{yuan2025autodriver} & Reflective & DAC & 95.9 & NAVSIM v1; navtest \\
Subscores & AutoVLA-RFT \citep{zhou2025autovla} & Compressed & DAC & 95.64 & NAVSIM v1; open-loop; test; 4s \\
Subscores & HiST-VLA \citep{wang2026histvla} & Tokenized & TTC & 99.4 & NAVSIM v2; navtest \\
Subscores & ReflectDrive$^\dagger$ \citep{li2025reflectdrive} & Reflective & TTC & 99.1 & NAVSIM v1; navtest \\
Subscores & ELF-VLA-8B \citep{luo2026unleashing} & Reflective & TTC & 98.4 & NAVSIM v2; navtest \\
Subscores & AutoVLA-RFT \citep{zhou2025autovla} & Compressed & TTC & 98.04 & NAVSIM v1; open-loop; test; 4s \\
Subscores & AutoVLA-BoN \citep{zhou2025autovla} & Compressed & TTC & 97.12 & NAVSIM v1; open-loop; test; 4s \\
Subscores & AutoDrive-P3-Detailed \citep{ye2026autodrivep3} & Compressed & TTC & 96.5 & NAVSIM v1; navtest \\
Subscores & ELF-VLA-8B \citep{luo2026unleashing} & Reflective & TTC & 96.0 & NAVSIM v1; navtest \\
Subscores & AutoDrive-R\textsuperscript{2}-7B \citep{yuan2025autodriver} & Reflective & TTC & 95.4 & NAVSIM v1; navtest \\
Subscores & AdaThinkDrive \citep{luo2025adathinkdrive} & Compressed & TTC & 95.2 & NAVSIM v1; navtest \\
Subscores & ReflectDrive \citep{li2025reflectdrive} & Reflective & TTC & 93.5 & NAVSIM v1; navtest \\
Subscores & AdaThinkDrive \citep{luo2025adathinkdrive} & Compressed & Comfort & 100 & NAVSIM v1; navtest \\
Subscores & AutoDrive-R\textsuperscript{2}-7B \citep{yuan2025autodriver} & Reflective & Comfort & 100 & NAVSIM v1; navtest \\
Subscores & AutoDrive-P3-Detailed \citep{ye2026autodrivep3} & Compressed & Comfort & 100 & NAVSIM v1; navtest \\
Subscores & ELF-VLA-8B \citep{luo2026unleashing} & Reflective & Comfort & 100 & NAVSIM v1; navtest \\
Subscores & ReflectDrive \citep{li2025reflectdrive} & Reflective & Comfort & 100 & NAVSIM v1; navtest \\
Subscores & AutoVLA-BoN \citep{zhou2025autovla} & Compressed & Comfort & 99.98 & NAVSIM v1; open-loop; test; 4s \\
Subscores & AutoVLA-RFT \citep{zhou2025autovla} & Compressed & Comfort & 99.94 & NAVSIM v1; open-loop; test; 4s \\
Subscores & ReflectDrive$^\dagger$ \citep{li2025reflectdrive} & Reflective & Comfort & 99.9 & NAVSIM v1; navtest \\
Subscores & HiST-VLA \citep{wang2026histvla} & Tokenized & DDC & 99.7 & NAVSIM v2; navtest \\
Subscores & ELF-VLA-8B \citep{luo2026unleashing} & Reflective & DDC & 99.4 & NAVSIM v2; navtest \\
Subscores & AutoVLA-BoN \citep{zhou2025autovla} & Compressed & DDC & 95.51 & NAVSIM v1; open-loop; test; 4s \\
Subscores & AutoVLA-RFT \citep{zhou2025autovla} & Compressed & DDC & 95.40 & NAVSIM v1; open-loop; test; 4s \\
Subscores & HiST-VLA \citep{wang2026histvla} & Tokenized & TLC & 99.9 & NAVSIM v2; navtest \\
Subscores & ELF-VLA-8B \citep{luo2026unleashing} & Reflective & TLC & 99.8 & NAVSIM v2; navtest \\
Subscores & HiST-VLA \citep{wang2026histvla} & Tokenized & LK & 98.9 & NAVSIM v2; navtest \\
Subscores & ELF-VLA-8B \citep{luo2026unleashing} & Reflective & LK & 96.9 & NAVSIM v2; navtest \\
Subscores & HiST-VLA \citep{wang2026histvla} & Tokenized & HC & 98.4 & NAVSIM v2; navtest \\
Subscores & ELF-VLA-8B \citep{luo2026unleashing} & Reflective & HC & 98.3 & NAVSIM v2; navtest \\
Subscores & ELF-VLA-8B \citep{luo2026unleashing} & Reflective & EC & 87.2 & NAVSIM v2; navtest \\
Subscores & HiST-VLA \citep{wang2026histvla} & Tokenized & EC & 66.0 & NAVSIM v2; navtest \\
Subscores & HiST-VLA \citep{wang2026histvla} & Tokenized & EP & 89.2 & NAVSIM v2; navtest \\
Subscores & ReflectDrive$^\dagger$ \citep{li2025reflectdrive} & Reflective & EP & 88.9 & NAVSIM v1; navtest \\
Subscores & ELF-VLA-8B \citep{luo2026unleashing} & Reflective & EP & 88.5 & NAVSIM v2; navtest \\
Subscores & AutoVLA-BoN \citep{zhou2025autovla} & Compressed & EP & 87.55 & NAVSIM v1; open-loop; test; 4s \\
Subscores & ReflectDrive \citep{li2025reflectdrive} & Reflective & EP & 86.9 & NAVSIM v1; navtest \\
Subscores & SGDrive \citep{li2026sgdrive} & Rollout-based & EP & 85.8 & NAVSIM v1; navtest RFT \\
Subscores & ELF-VLA-8B \citep{luo2026unleashing} & Reflective & EP & 85.3 & NAVSIM v1; navtest \\
Subscores & AutoDrive-P3-Detailed \citep{ye2026autodrivep3} & Compressed & EP & 84.8 & NAVSIM v1; navtest \\
Subscores & AdaThinkDrive \citep{luo2025adathinkdrive} & Compressed & EP & 84.4 & NAVSIM v1; navtest \\
Subscores & AutoDrive-R\textsuperscript{2}-7B \citep{yuan2025autodriver} & Reflective & EP & 82.7 & NAVSIM v1; navtest \\
Subscores & AutoVLA-RFT \citep{zhou2025autovla} & Compressed & EP & 81.87 & NAVSIM v1; open-loop; test; 4s \\
\midrule
\multicolumn{6}{l}{\textbf{Safety / latency / auxiliary}} \\
\midrule
Safety / latency / auxiliary & AutoVLA-BoN \citep{zhou2025autovla} & Compressed & Collision score & 99.14 & NAVSIM; open-loop; test; 4s \\
Safety / latency / auxiliary & AutoVLA-RFT \citep{zhou2025autovla} & Compressed & Collision score & 98.41 & NAVSIM; open-loop; test; 4s \\
Safety / latency / auxiliary & DiffVLA \citep{jiang2025diffvla} & Optimization-based & DAC (stage 2) & 88.84 & NAVSIM v2; AGC private test; stage 2 \\
Safety / latency / auxiliary & DiffVLA \citep{jiang2025diffvla} & Optimization-based & NC (stage 2) & 81.27 & NAVSIM v2; AGC private test; stage 2 \\
Safety / latency / auxiliary & DiffVLA \citep{jiang2025diffvla} & Optimization-based & TTC (stage 2) & 76.46 & NAVSIM v2; AGC private test; stage 2 \\
Safety / latency / auxiliary & ELF-VLA \citep{luo2026unleashing} & Reflective & Path Acc. & 92.5 & NAVSIM; high-level planning \\
Safety / latency / auxiliary & OneVL \citep{team2026onevl} & Tokenized & PDM-score & 88.84 & NAVSIM; planning benchmark; test \\
Safety / latency / auxiliary & ELF-VLA \citep{luo2026unleashing} & Reflective & Accuracy & 80.3 & NAVSIM; high-level planning \\
Safety / latency / auxiliary & OneVL \citep{team2026onevl} & Tokenized & Meta Action Accuracy & 71.00 & NAVSIM; text CoT quality; test \\
Safety / latency / auxiliary & Reasoning-VLA-7B \citep{zhang2025reasoningvla} & Compressed & L2 avg & 0.22 & NAVSIM; generalized open-loop; recommended split; 3s \\
Safety / latency / auxiliary & OneVL \citep{team2026onevl} & Tokenized & Latency (MLP head) & 0.24 & NAVSIM; real-world deployment; test \\
Safety / latency / auxiliary & AdaThinkDrive \citep{luo2025adathinkdrive} & Compressed & Inference time (s) & 0.74 & NAVSIM; closed-loop; test; 4s \\
Safety / latency / auxiliary & OneVL \citep{team2026onevl} & Tokenized & Latency (full pipeline) & 4.46 & NAVSIM; planning benchmark; test \\
Safety / latency / auxiliary & DriveDreamer-Policy \citep{zhou2026drivedreamerpolicy} & Predictive & AbsRel & 8.1 & NAVSIM; depth generation; test \\
Safety / latency / auxiliary & DriveDreamer-Policy \citep{zhou2026drivedreamerpolicy} & Predictive & FVD & 53.59 & NAVSIM; world generation; test \\
\end{longtable}
\normalsize

\subsection*{Bench2Drive / CARLA}

\scriptsize
\setlength{\tabcolsep}{1pt}
\renewcommand{\arraystretch}{1.08}

\begin{longtable}{>{\raggedright\arraybackslash}p{0.09\textwidth} >{\raggedright\arraybackslash}p{0.16\textwidth} >{\raggedright\arraybackslash}p{0.10\textwidth} >{\raggedright\arraybackslash}p{0.22\textwidth} >{\raggedright\arraybackslash}p{0.055\textwidth} >{\raggedright\arraybackslash}p{0.29\textwidth}}
\caption{Reported results for Bench2Drive / CARLA grouped by metric type.}
\label{tab:reported-bench2drive-carla} \\
\toprule
Metric group & Method & Representation & Metric & Value & Protocol / Notes \\
\midrule
\endfirsthead
\toprule
Metric group & Method & Representation & Metric & Value & Protocol / Notes \\
\midrule
\endhead
\midrule \multicolumn{6}{r}{Continued on next page} \\
\endfoot
\bottomrule
\endlastfoot

\midrule
\multicolumn{6}{l}{\textbf{Driving performance}} \\
\midrule
Driving performance & AutoVLA \citep{zhou2025autovla} & Compressed & Driving Score (0--100) & 78.84 & Bench2Drive; closed-loop; test \\
Driving performance & MindDrive \citep{fu2025minddrive} & Compressed & Driving Score (0--100) & 78.04 & Bench2Drive; closed-loop \\
Driving performance & SpaceDrive+ \citep{li2025spacedrive} & Action-grounded & Driving Score (0--100) & 78.02 & Bench2Drive; closed-loop \\
Driving performance & ORION \citep{fu2025orion} & Descriptive & Driving Score (0--100) & 77.74 & Bench2Drive; closed-loop \\
Driving performance & ReCogDrive \citep{li2025recogdrive} & Optimization-based & Driving Score (0--100) & 71.36 & Bench2Drive; closed-loop \\
Driving performance & MindDriver \citep{zhang2026minddriver} & Action-grounded & Driving Score (0--100) & 65.48 & Bench2Drive; closed-loop \\
Driving performance & LeAD \citep{zhang2025lead} & Knowledge & Driving Score (0--100) & 71.96 & CARLA; Leaderboard V1; closed-loop \\
Driving performance & LangCoop \citep{gao2025langcoop} & Cooperative & Driving Score (0--100) & 48.8 & CARLA; closed-loop \\
Driving performance & AutoVLA \citep{zhou2025autovla} & Compressed & Success Rate (0--100) & 57.73 & Bench2Drive; closed-loop; test \\
Driving performance & SpaceDrive+ \citep{li2025spacedrive} & Action-grounded & Success Rate (0--100) & 55.11 & Bench2Drive; closed-loop \\
Driving performance & MindDrive \citep{fu2025minddrive} & Compressed & Success Rate (0--100) & 55.09 & Bench2Drive; closed-loop \\
Driving performance & ORION \citep{fu2025orion} & Descriptive & Success Rate (0--100) & 54.62 & Bench2Drive; closed-loop \\
Driving performance & ReCogDrive \citep{li2025recogdrive} & Optimization-based & Success Rate (0--100) & 45.45 & Bench2Drive; closed-loop \\
Driving performance & MindDriver \citep{zhang2026minddriver} & Action-grounded & Success Rate (0--100) & 39.55 & Bench2Drive; closed-loop \\
Driving performance & DriveMind \citep{wasif2025drivemind} & Compressed & Route Completion (0--1) & 0.98 & CARLA; closed-loop; Town02 \\
Driving performance & DriveMind \citep{wasif2025drivemind} & Compressed & Success Rate (0--1) & 0.97 & CARLA; closed-loop; Town02 \\
Driving performance & COVLM-RL-Town03 \citep{li2025covlmrl} & Compressed & Route Completion (0--1) & 0.88 & CARLA; closed-loop; unseen Town03 \\
Driving performance & COVLM-RL-Town02 \citep{li2025covlmrl} & Compressed & Route Completion (0--1) & 0.78 & CARLA; closed-loop; trained Town02 \\
Driving performance & COVLM-RL-Town03 \citep{li2025covlmrl} & Compressed & Success Rate (0--1) & 0.70 & CARLA; closed-loop; unseen Town03 \\
Driving performance & COVLM-RL-Town02 \citep{li2025covlmrl} & Compressed & Success Rate (0--1) & 0.70 & CARLA; closed-loop; trained Town02 \\
Driving performance & COVLM-RL-Town03 \citep{li2025covlmrl} & Compressed & CDS (0--1, $\downarrow$) & 0.64 & CARLA; closed-loop; unseen Town03 \\
Driving performance & COVLM-RL-Town02 \citep{li2025covlmrl} & Compressed & CDS (0--1, $\downarrow$) & 0.63 & CARLA; closed-loop; trained Town02 \\
Driving performance & CoLMDriver \citep{liu2025colmdriver} & Cooperative & Driving Score (0--100) & 88.53 & CARLA; InterDrive-total; closed-loop \\
Driving performance & WiseAD \citep{zhang2024wisead} & Descriptive & Driving Score (relative, \%) & +11.9 & CARLA; closed-loop; improvement over baseline \\
Driving performance & SteerVLA \citep{gao2026steervlab} & Descriptive & Driving Score (relative, pts) & +4.77 & Bench2Drive; improvement over prior SOTA \\
Driving performance & DriveMLM \citep{wang2023drivemlm} & Procedural & Driving Score (relative, pts) & +4.7 & CARLA Town05 Long; improvement over Apollo baseline \\\midrule
\multicolumn{6}{l}{\textbf{Safety / infractions}} \\
\midrule
Safety / infractions & DriveMind \citep{wasif2025drivemind} & Compressed & Collision speed (km/h) & 0.01 & CARLA; closed-loop; held-out maps (Towns 1/3/4/5) \\
Safety / infractions & UniDrive-WM \citep{xiong2026unidrivewm} & Predictive & Collision rate reduction (relative, \%) & 10.4 & Bench2Drive; closed-loop; test \\
\midrule
\multicolumn{6}{l}{\textbf{Auxiliary}} \\
\midrule
Auxiliary & MindDrive \citep{fu2025minddrive} & Compressed & Overtaking ability & 75.56 & Bench2Drive; closed-loop; official benchmark \\
Auxiliary & MindDrive \citep{fu2025minddrive} & Compressed & Emergency Brake ability & 68.33 & Bench2Drive; closed-loop; official benchmark \\
Auxiliary & MindDrive \citep{fu2025minddrive} & Compressed & Traffic Sign ability & 57.89 & Bench2Drive; closed-loop; official benchmark \\
Auxiliary & MindDrive \citep{fu2025minddrive} & Compressed & Ability mean & 56.94 & Bench2Drive; closed-loop; official benchmark \\
Auxiliary & MindDrive \citep{fu2025minddrive} & Compressed & Give Way ability & 50.00 & Bench2Drive; closed-loop; official benchmark \\
Auxiliary & MindDrive \citep{fu2025minddrive} & Compressed & Merging ability & 32.89 & Bench2Drive; closed-loop; official benchmark \\
Auxiliary & AutoVLA \citep{zhou2025autovla} & Compressed & Efficiency & 146.93 & Bench2Drive; closed-loop; test \\
Auxiliary & AutoVLA \citep{zhou2025autovla} & Compressed & Comfortness & 39.33 & Bench2Drive; closed-loop; test \\
Auxiliary & COVLM-RL-Town02 \citep{li2025covlmrl} & Compressed & Traveled Distance (TD) & 1458.95 & CARLA; closed-loop; Town02 (trained) \\
Auxiliary & COVLM-RL-Town03 \citep{li2025covlmrl} & Compressed & Traveled Distance (TD) & 3468.47 & CARLA; closed-loop; Town03 (unseen) \\
Auxiliary & COVLM-RL-Town02 \citep{li2025covlmrl} & Compressed & Average Distance (AD) & 145.89 & CARLA; closed-loop; Town02 (trained) \\
Auxiliary & COVLM-RL-Town03 \citep{li2025covlmrl} & Compressed & Average Distance (AD) & 346.96 & CARLA; closed-loop; Town03 (unseen) \\
Auxiliary & COVLM-RL-Town02 \citep{li2025covlmrl} & Compressed & Speed Mean (SM) & 18.74 & CARLA; closed-loop; Town02 (trained) \\
Auxiliary & COVLM-RL-Town03 \citep{li2025covlmrl} & Compressed & Speed Mean (SM) & 21.38 & CARLA; closed-loop; Town03 (unseen) \\
Auxiliary & COVLM-RL-Town02 \citep{li2025covlmrl} & Compressed & Speed Std (SS) & 3.34 & CARLA; closed-loop; Town02 (trained) \\
Auxiliary & COVLM-RL-Town03 \citep{li2025covlmrl} & Compressed & Speed Std (SS) & 3.25 & CARLA; closed-loop; Town03 (unseen) \\
Auxiliary & COVLM-RL-Town02 \citep{li2025covlmrl} & Compressed & Reward Mean (RM) & 4.29 & CARLA; closed-loop; Town02 (trained) \\
Auxiliary & COVLM-RL-Town03 \citep{li2025covlmrl} & Compressed & Reward Mean (RM) & 4.41 & CARLA; closed-loop; Town03 (unseen) \\
Auxiliary & COVLM-RL-Town02 \citep{li2025covlmrl} & Compressed & Reward Std (RS) & 1.20 & CARLA; closed-loop; Town02 (trained) \\
Auxiliary & COVLM-RL-Town03 \citep{li2025covlmrl} & Compressed & Reward Std (RS) & 1.33 & CARLA; closed-loop; Town03 (unseen) \\
Auxiliary & COVLM-RL-Town02 \citep{li2025covlmrl} & Compressed & Centerline Deviation Mean (CDM) & 0.71 & CARLA; closed-loop; Town02 (trained) \\
Auxiliary & COVLM-RL-Town03 \citep{li2025covlmrl} & Compressed & Centerline Deviation Mean (CDM) & 0.82 & CARLA; closed-loop; Town03 (unseen) \\
Auxiliary & DriveMind \citep{wasif2025drivemind} & Compressed & Total distance (m) & 2083.10 & CARLA; closed-loop; held-out maps \\
Auxiliary & DriveMind \citep{wasif2025drivemind} & Compressed & Average speed (km/h) & 19.40 & CARLA; closed-loop; held-out maps \\
Auxiliary & AFSP \citep{chen2026bridging} & Compressed & Average Lateral Deviation (m) & 1.21 & CARLA; closed-loop; Scenario 1 (nominal map) \\
Auxiliary & AFSP \citep{chen2026bridging} & Compressed & Average Lateral Deviation (m) & 1.01 & CARLA; closed-loop; Scenario 2 (shifted map) \\
Auxiliary & AFSP \citep{chen2026bridging} & Compressed & Average Lateral Deviation (m) & 1.02 & CARLA; closed-loop; Scenario 3 (shifted map) \\
Auxiliary & AFSP \citep{chen2026bridging} & Compressed & Maximum Lateral Deviation (m) & 3.72 & CARLA; closed-loop; Scenario 1 (nominal map) \\
Auxiliary & AFSP \citep{chen2026bridging} & Compressed & Maximum Lateral Deviation (m) & 3.23 & CARLA; closed-loop; Scenario 2 (shifted map) \\
Auxiliary & AFSP \citep{chen2026bridging} & Compressed & Maximum Lateral Deviation (m) & 3.42 & CARLA; closed-loop; Scenario 3 (shifted map) \\
Auxiliary & AFSP \citep{chen2026bridging} & Compressed & Speed Variation (m/s) & 1.56 & CARLA; closed-loop; Scenario 1 (nominal map) \\
Auxiliary & AFSP \citep{chen2026bridging} & Compressed & Speed Variation (m/s) & 1.30 & CARLA; closed-loop; Scenario 2 (shifted map) \\
Auxiliary & AFSP \citep{chen2026bridging} & Compressed & Speed Variation (m/s) & 1.48 & CARLA; closed-loop; Scenario 3 (shifted map) \\
Auxiliary & AFSP \citep{chen2026bridging} & Compressed & Finish Time (s) & 14.85 & CARLA; closed-loop; Scenario 1 (nominal map) \\
Auxiliary & AFSP \citep{chen2026bridging} & Compressed & Finish Time (s) & 15.02 & CARLA; closed-loop; Scenario 2 (shifted map) \\
Auxiliary & AFSP \citep{chen2026bridging} & Compressed & Finish Time (s) & 14.73 & CARLA; closed-loop; Scenario 3 (shifted map) \\
Auxiliary & AFSP \citep{chen2026bridging} & Compressed & Trajectory Length (m) & 84.21 & CARLA; closed-loop; Scenario 1 (nominal map) \\
Auxiliary & AFSP \citep{chen2026bridging} & Compressed & Trajectory Length (m) & 83.86 & CARLA; closed-loop; Scenario 2 (shifted map) \\
Auxiliary & AFSP \citep{chen2026bridging} & Compressed & Trajectory Length (m) & 83.72 & CARLA; closed-loop; Scenario 3 (shifted map) \\
Auxiliary & ReasonPlan \citep{liu2025reasonplan} & Procedural & L2 error reduction (relative, \%) & 19.0 & Bench2Drive; open-loop planning; vs. E2E imitation baseline \\
Auxiliary & UniDrive-WM \citep{xiong2026unidrivewm} & Predictive & L2 trajectory error reduction (relative, \%) & 7.3 & Bench2Drive; open-loop planning; test \\
\end{longtable}
\normalsize

\end{document}